\documentclass[10pt,journal,compsoc]{IEEEtran}

\usepackage{stmaryrd}
\usepackage{url}
\usepackage{color}
\usepackage[export]{adjustbox}
\usepackage{xspace}
\usepackage{multirow}
\usepackage{dingbat}
\usepackage{csquotes}
\usepackage[table]{xcolor}
\usepackage{epigraph}
\usepackage{amsmath}

\usepackage{xspace}
\newcommand{\latinphrase}[1]{\textit{#1}} 
\newcommand{\etal}{\latinphrase{et~al.}\xspace}
\newcommand{\ie}{\latinphrase{i.e.}\xspace}

\newcommand{\eg}{\latinphrase{e.g.}\xspace}
\newcommand{\etc}{\latinphrase{etc.}\xspace}

\ifCLASSOPTIONcompsoc
  \usepackage[nocompress]{cite}
\else
  \usepackage{cite}
\fi

\ifCLASSINFOpdf
\else
\fi

\begin{document}
\title{A Deep Journey into Super-resolution: A Survey}

\author{Saeed~Anwar, 
        Salman~Khan, 
        and~Nick~Barnes 
\IEEEcompsocitemizethanks{\IEEEcompsocthanksitem Saeed Anwar is with Data61, CSIRO, Australia.\protect\\
E-mail: saeed.anwar@csiro.au
\IEEEcompsocthanksitem Salman Khan is with IIAI, UAE and ANU, Australia.
\IEEEcompsocthanksitem Nick Barnes is with Data61, CSIRO, Australia.}
}


\def\NB#1{{\color{red} {{#1}}}} %

\IEEEtitleabstractindextext{%
\begin{abstract}
Deep convolutional networks based super-resolution is a fast-growing field with numerous practical applications. In this exposition, we extensively compare more than 30 state-of-the-art super-resolution Convolutional Neural Networks (CNNs) over three classical and three recently introduced challenging datasets to benchmark single image super-resolution.  We introduce a taxonomy for deep-learning based super-resolution networks that groups existing methods into nine categories including linear, residual, multi-branch, recursive, progressive, attention-based and adversarial designs. We also provide comparisons between the models in terms of network complexity, memory footprint, model input and output, learning details, the type of network losses and important architectural differences (\eg, depth, skip-connections, filters). The extensive evaluation performed, shows the consistent and rapid growth in the accuracy in the past few years along with a corresponding boost in model complexity and the availability of large-scale datasets. It is also observed that the pioneering methods identified as the benchmark have been significantly outperformed by the current contenders.  Despite the progress in recent years, we identify several shortcomings of existing techniques and provide future research directions towards the solution of these open problems. Datasets and Codes for evaluation are made publicly available at \url{https://github.com/saeed-anwar/SRsurvey}.
\end{abstract}


}

\maketitle

\IEEEdisplaynontitleabstractindextext

\IEEEpeerreviewmaketitle

\IEEEraisesectionheading{\section{Introduction}\label{sec:introduction}}

\setlength{\epigraphwidth}{23em}
\epigraph{\it `Everything  has been said before, but since nobody listens
we have to keep going back and beginning all over again.'}{Andre Gide}

\IEEEPARstart{I}{mage} super-resolution (SR) 
has received increasing attention from {the} research community in recent years.  Super-resolution aims to convert a given low-resolution image with coarse details to a corresponding high-resolution image with better visual quality and refined details. Image super-resolution is also
{referred to} by other names such as image scaling, interpolation, upsampling, zooming and enlargement. The process of generating a raster image with higher resolution can be performed using a single image or multiple images. Due to practical considerations, this exposition mainly focuses on single image super-resolution (SISR) which has been extensively studied due to its challenging nature. For SR in higher dimensional inputs (such as videos and 3D scans), we refer the reader to recent seminal works \cite{jo2018deep,wang2019edvr,li2019fast,thawakar2019image,smith2018multi}. 

High-resolution images provide improved reconstructed details of the scenes and constituent objects, which are critical for many devices such as large computer displays, HD television sets, and hand-held devices (mobile phones, tablets, cameras \etc). Furthermore, super-resolution has important applications in many other domains \eg object detection in scenes~\cite{girshick2016region} (particularly small {objects} \cite{bai2018sod}), face recognition in surveillance videos~\cite{mudunuri2016low}, medical imaging~\cite{greenspan2008super},  improving interpretation of images in remote sensing \cite{lillesand2014remote}, astronomical images \cite{lobanov2005resolution},  and forensics \cite{swaminathan2008digital}.

Super-resolution is a classical problem that is still considered a challenging and open research problem in computer vision due to several reasons. Firstly, SR is an ill-posed inverse problem, \ie an under-determined case. Instead of a single unique solution, there exist multiple solutions for the same low-resolution image. To constrain the solution-space, reliable prior information is typically required.  Secondly, the complexity of the problem increases as the up-scaling factor increases. At higher factors, the recovery of missing scene details becomes even more complex, and consequently it often leads to reproduction of wrong information. Furthermore, 
assessment of the quality of output
is not straightforward \ie, 
quantitative metrics (\eg PSNR, SSIM) only loosely correlate to human perception.

Super-resolution methods can be broadly divided into two main categories: traditional and deep learning methods. Classical algorithms have been around for decades now, but are out-performed by their deep learning based counterparts. Therefore, most recent algorithms rely on data-driven deep learning models to reconstruct the required details for accurate super-resolution. Deep learning is a branch of machine learning, 
{that} aims to automatically learn the relationship between input and output directly from the data. Alongside SR, deep learning algorithms have shown promising results on other sub-fields in Artificial Intelligence \cite{khan2018guide} such as object classification \cite{he2016deep} and detection~\cite{ren2015faster}, natural language processing~\cite{kumar2016ask,collobert2011natural}, image processing~\cite{anwar2018deep,anwar2017chaining}, and audio signal processing~\cite{dahl2012context}. Due to these reasons, in this survey, we mainly focus on deep learning algorithms for SR and only provide a brief background on traditional approaches (Section \ref{sec:background}). 

\textbf{Our Contributions:} In this exposition, our focus is on deep neural networks for single (natural) image super-resolution. Our contribution is five-fold. 1) We provide a thorough review of the recent techniques 
{for} image super-resolution. 2) We introduce a new taxonomy of the SR algorithms based on their structural differences. 3) A comprehensive analysis is performed based on the number of parameters,  algorithm settings, training details and important architectural innovations that leads to significant performance improvements. 4) We provide a systematic evaluation of algorithms on six publicly available datasets for SISR. 5) We discuss the challenges and provide insights into the possible future directions. 

\section{Background}\label{sec:background}
Let us consider a Low-Resolution (LR) image is denoted by $\bf{y}$ and the corresponding high-resolution (HR) image is denoted by $\bf x$, then the degradation process is given as:
\begin{equation}
\bf{y} = \Phi(\bf{x};\theta_{\eta} ),
\label{eq:degradation_process}
\end{equation}
where $\Phi$ is the degradation function, and $\theta_{\eta}$ denotes the degradation  parameters (such as the scaling factor, noise \etc). In a real-world scenario, only $\mathbf{y}$ is available while no information about the degradation process or the degradation parameters $\theta_{\eta}$. Super-resolution seeks to nullify the degradation effect and recovers an approximation $\hat{\mathbf{x}}$ of the ground-truth image $\mathbf{x}$ as,
\begin{equation}
\hat{\bf{x}} = \Phi^{-1}(\bf{y}, \theta_{\varsigma}),
\label{eq:SR}
\end{equation}
where, $\theta_{\varsigma}$ are the parameters for the function $\Phi^{-1}$. The degradation process is unknown and can be quite complex. It can be affected by several factors such as noise (sensor and speckle), compression, blur (defocus and motion), and other artifacts. Therefore, most research works prefer the following degradation model over that of Eq.~\ref{eq:degradation_process}:
\begin{equation}
\mathbf{y} = (\mathbf{x} \otimes \mathbf{k}) \downarrow_{s}~+~\mathbf{n},
\label{eq:SR_down}
\end{equation}
where $\mathbf{k}$ is the blurring kernel and $\mathbf{x} \otimes \mathbf{k}$ is the convolution operation between the HR image and the blur kernel, $\downarrow$ is a downsampling operation with a  scaling factor $s$. The variable $\mathbf{n}$ denotes the additive white Gaussian noise (AWGN) with a standard deviation of $\sigma$ (noise level). In image super-resolution, the aim is to minimize the data fidelity term associated with the model $\mathbf{y} = \mathbf{x} \otimes \mathbf{k} + \mathbf{n}$, as,
\begin{equation}
J(\mathbf{\hat{x}},\theta_{\varsigma}, \mathbf{k}) = \underbrace{\|\mathbf{x} \otimes \mathbf{k} - \mathbf{y}\|}_{\text{data fidelity term}} + \alpha \underbrace{\Psi(\mathbf{x},\theta_{\varsigma})}_{\text{regularizer}},
\label{eq:cost_func}  
\end{equation}
where $\alpha$ is the balancing factor for the the data fidelity term and image prior  $\Psi(\cdot)$. According to Yang~\etal~\cite{yang2014single}, based on the image prior, super-resolution methods can be roughly categorized into: prediction methods~\cite{irani1991improving}, edge-based methods~\cite{fattal2007image}, statistical methods~\cite{huang1999statistics}, patch-based methods~\cite{freeman2002example,chang2004super,yang2013fast}, and deep learning methods~\cite{dong2016SRCNNPAMI}. In this article, our focus is on the methods which employ deep neural networks to learn the prior. 

\begin{figure*}
\centering
\includegraphics[width=\textwidth,trim={0.1cm 13cm 4cm 1cm},clip]{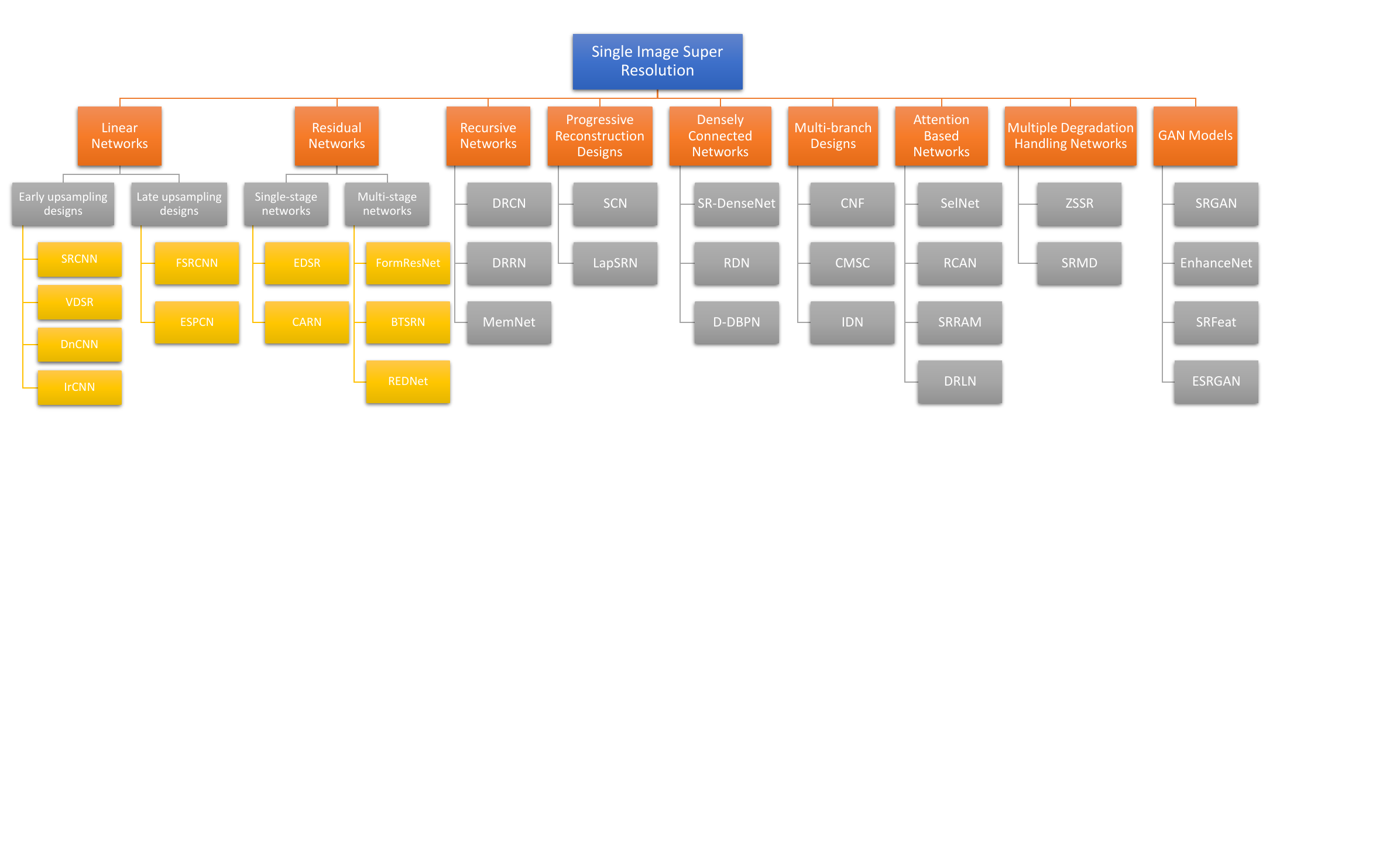}
\caption{The taxonomy of the existing single-image super-resolution techniques based on the most distinguishing features.}
\label{fig:overall}
\end{figure*}

\section{Single Image Super-resolution}
\label{sec:sisr}
The SISR problem has been extensively studied in the literature using a variety of deep learning based techniques. We categorize 
existing methods into nine groups according to the most distinctive features in their model designs. The overall taxonomy used in this literature is shown in Figure~\ref{fig:overall}. Among these, we begin discussion with the earliest and simplest network designs that are called the linear networks.  

\subsection{Linear networks}
Linear networks have a simple structure consisting of only a single path for signal flow without any skip connections or multiple-branches. In such network designs, several convolution layers are stacked on top of each other and the input flows sequentially from initial to later layers. 
Linear networks differ in the way the up-sampling operation is performed \ie, early upsampling or late upsampling. Note that some 
linear networks learn to reproduce the residual image \ie, the difference between the LR and HR images \cite{kim2016VDSR, zhang2017DnCNN, zhang2017IrCNN}. Since the network architecture is linear in such cases, we categorize them as linear networks.
{This is as opposed to} residual networks that have s{k}ip connections in their design (Sec.~\ref{sec:residual}). We elaborate notable linear network designs in these two sub-categories below.

\subsubsection{Early Upsampling Designs}
The early upsampling designs are 
linear networks that first upsample the LR input to match with desired HR output size and then learn hierarchical feature representations to generate the output. A common upsampling operation used for this purpose is 
Bicubic interpolation, which is a computationally expensive operation. A seminal work based on this pipeline is the SRCNN which we explain next. 

\noindent
$\bullet$ \textbf{SRCNN:}
Super-Resolution Convolutional Neural Network abbreviated as SRCNN~\cite{dong2014SRCNN,dong2016SRCNNPAMI} is the first successful attempt towards using only convolutional layers for super-resolution. This effort can rightfully be considered as the pioneering work in deep learning based SR that inspired several later attempts in this direction. SRCNN structure is straightforward, it only consists of convolutional layers where each layer (except the last one) is followed by rectified linear unit (ReLU) non-linearity. There are a total of three convolutional and two ReLU layers, stacked together linearly. Although the layers are the same (\ie, convolution layers),
 the authors named the layers according to their functionality. The \emph{first} convolutional layer is termed as {patch extraction} or feature extraction which creates the feature maps from the input images. The \emph{second} convolutional layer is called {non-linear mapping} which converts the feature maps onto
 high-dimensional feature vectors. The \emph{last} convolutional layer aggregates the features maps to output the final high-resolution image. The structure of SRCNN is shown in the Figure~\ref{fig:archs}. 

The training data set is synthesized by extracting non-overlapping dense patches of size 32$\times$32 from the HR images. The LR input patches are first downsampled and then upsampled using bicubic interpolation having the same size as the high-resolution output image. The SRCNN is an end-to-end trainable network and minimizes the difference between the output reconstructed high-resolution images and the ground truth high-resolution images using Mean Squared Error (MSE) loss function.

\noindent
$\bullet$\textbf{VDSR:}
Unlike the shallow network architectures used in SRCNN \cite{dong2016SRCNNPAMI} and FSRCNN \cite{dong2016FSRCNN},  Very Deep Super-Resolution \cite{kim2016VDSR} (VDSR) is based on a deep CNN architecture originally proposed in \cite{simonyan2014very}. This architecture is popularly known as the VGG-net and uses fixed-size convolutions ($3{\times}3$) in all network layers. To avoid slow convergence in deep networks (specifically with 20 weight layers), they propose two effective strategies. Firstly, instead of directly generating a HR image, they learn a residual mapping that generates the difference between the HR and LR image. As a result, it provides an easier objective and the network focuses on only high-frequency information. Secondly, gradients are clipped with in the range $[-\theta, +\theta]$ which allows very high learning rates to speed up the training process. Their results support the argument that deeper networks can provide better contextualization and learn generalizable representations that can be used for multi-scale super-resolution.

\noindent
$\bullet$ \textbf{DnCNN}
~\cite{zhang2017DnCNN} learns to predict a high-frequency residual {directly} instead of the latent super-resolved image.
The residual image is basically the difference between LR and HR images. The architecture of DnCNN is very simple and similar to SRCNN as it only stacks 
convolutional, batch normalization and ReLU layers. The architecture of DnCNN is shown in Figure \ref{fig:archs}.

Although both models were able to report favorable results, their performance depends heavily on the accuracy of noise estimation without knowing the underlying structures and textures present in the image. Besides, they are computationally expensive because of the batch normalization operations after every convolutional layer. 

\noindent
$\bullet$ \textbf{IRCNN:}
Image Restoration CNN (IRCNN) \cite{zhang2017IrCNN} proposes a set of CNN based denoisers that can be jointly used for several low-level vision tasks such as image denoising, deblurring and super-resolution. This technique aims to combine high-performing discriminative CNN networks with model-based optimization approaches to achieve better generalizability across image restoration tasks. Specifically, the Half Quadratric Splitting (HQS) technique is used to uncouple regularization and fidelity terms in the observation model~\cite{geman1995nonlinear} . Afterwards, a denoising prior is discriminatively learned using a CNN due to its superior modeling capacity and test time efficiency. The CNN denoiser is composed of a stack of 7 dilated convolution layers interleaved with batch normalization and ReLU non-linearity layers. The dilation operation helps in modeling larger context by enclosing a bigger receptive field. To speed up the learning process, residual image learning is performed {in a} similar {manner} to previous architectures such as VDSR \cite{kim2016VDSR}, DRCN \cite{kim2016DRCN} and DRRN \cite{tai2017DRRN}. {The }authors also proposed to use small sized training samples along with zero-padding to avoid boundary artifacts due to {the} convolution operation. 

A set of 25 denoisers is trained with the range of noise levels [0,50] that are collectively used for image restoration tasks. The proposed unified approach provides strong performance simultaneously on image denoising, deblurring and super-resolution.

\subsubsection{Late Upsampling Designs}
As we saw in the previous examples, linear networks generally perform early upsampling on the input images. This operation can be computationally expensive since the later network structure grows in proportion to deal with larger sized inputs. To address this problem, post-upsampling networks perform learning on the low-resolution inputs and then upsample the features near the output of the network. This strategy results in efficient approaches with low memory footprint. We discuss such designs in the following.

\noindent
$\bullet$ \textbf{FSRCNN:}
Fast Super-Resolution Convolutional Neural Network (FSRCNN)~\cite{dong2016FSRCNN} 
{improves} speed and quality over SRCNN~\cite{dong2014SRCNN}. The aim is to bring the rate of computation to real-time (24 fps) as compared to SRCNN (1.3 fps).  FSRCNN~\cite{dong2016FSRCNN} also has a simple architecture and consists of four convolution layers and one deconvolution. The architecture of FSRCNN~\cite{dong2016FSRCNN} is shown in Figure~\ref{fig:archs}.

Although the first four layers implement convolution operations, FSRCNN~\cite{dong2016FSRCNN} names each layer according to its function, namely \ie \emph{feature extraction}, \emph{shrinking},  \emph{non-linear mapping}, and \emph{expansion} layers.  The feature extraction step is similar to SRCNN~\cite{dong2014SRCNN}, 
the only difference lies in the input size and the filter size. The input to SRCNN~\cite{dong2014SRCNN} is an upsampled bicubic patch while the input to FSRCNN~\cite{dong2016FSRCNN} is the original patch without upsampling it.  The second convolution layer is named shrinking layer due to its ability to reduce the feature dimensions (number of parameters) by adopting a smaller filter size (\ie f=1) to increase computational efficiency.  Next, the convolutional layer acts as a non-linear mapping step, and according to the authors, this is a critical step both in SRCNN \cite{dong2014SRCNN} and FSRCNN~\cite{dong2016FSRCNN},  as it helps in learning non-linear functions and consequently has a strong influence on the performance. Through experimentation, the size of filters in the non-linear mapping layer is set to three, while the number of channels is kept the same as the previous layer.  The last convolutional layer, termed as expanding, is an inverse operation of the shrinking step to increase the number of dimensions. This layer results in an increase in performance by 0.3dB.

The final part of the network is an upsampling and aggregating deconvolution layer, which is an inverse process of the convolution.  In convolution operation, the image is convolved with the convolution filter with a stride, and the output of that convolutional layer is 1/stride of the input. However, the role of the filter is exactly opposite in deconvolutional layer, and here stride acts as an upscaling factor.  Similarly, another subtle difference from SRCNN~\cite{dong2014SRCNN} is the usage of Parametric Rectified Linear Unit (PReLU) \cite{he2015PReLU} instead of the Rectified Linear Unit (ReLU) after each convolutional layer.

FSRCNN\cite{dong2016FSRCNN} employs the same cost function as SRCNN~\cite{dong2014SRCNN} \ie mean-square error. For training, \cite{dong2016FSRCNN} used the 91-image dataset~\cite{yang2010image91} with another 100 images collected from the internet. Data augmentation such as rotation, flipping, and scaling is also employed to increase the number of images by 19 times. 

\noindent
$\bullet$ \textbf{ESPCN:}
Efficient sub-pixel convolutional neural network (ESPCN) \cite{shi2016ESPCN} is a fast SR approach that can operate in real-time both for images and videos.  As discussed above, traditional SR techniques first map the LR image to higher resolution usually with bi-cubic interpolation and subsequently learn the SR model in the higher dimensional space. ESPCN noted that this pipeline results in much higher computational requirements and alternatively propose to perform feature extraction in the LR space. After the features are extracted, ESPCN uses a sub-pixel convolution layer at the very end to aggregate LR feature maps and simultaneously perform projection to high dimensional space to reconstruct {the} HR image. Feature processing in LR space significantly reduces the memory and computational requirements.  

The sub-pixel convolution operation used in this work is essentially similar to convolution transpose or deconvolution operation \cite{shi2016deconvolution}, where a fractional kernel  stride is used to increase the spatial resolution of input feature maps. A separate upscaling kernel is used to map each feature map that provides more flexibility in modeling the LR to HR mapping. An $\ell_1$ loss is used to train the overall network. ESPCN provides 
competitive SR performance with efficiency as high as real-time processing of 1080p videos on  a single GPU.

\subsection{Residual Networks}\label{sec:residual}
In contrast to linear networks, residual learning uses skip connections in the network design to avoid gradients vanishing and makes it feasible to design very deep networks. Its significance was first demonstrated for {the} image classification problem \cite{he2016deep}. Recently, several networks~\cite{lim2017EDSR,ahn2018CARN} provided a boost to 
SR performance using 
residual learning. In this approach, 
algorithms learn residue \ie the high-frequencies between the input and ground-truth. Based on the number of stages used in such networks, we categorize existing residual learning approaches into single-stage \cite{lim2017EDSR, ahn2018CARN} and multi-stage networks \cite{jiao2017formresnet, Fan2017BTSRN, mao2016REDNet}.


\subsubsection{Single-stage Residual Nets}
A single-stage design is composed of a single network; examples are shown next.

\noindent
$\bullet$ \textbf{EDSR:}
The Enhanced Deep Super-Resolution (EDSR) \cite{lim2017EDSR} modifies the ResNet architecture \cite{he2016deep} proposed originally for image classification to work with the SR task. Specifically, they demonstrated substantial improvements by removing Batch Normalization layers (from each residual block) and ReLU activation (outside residual blocks). Similar to VDSR, they also extended their single scale approach to work on multiple scales. Their proposed Multi-scale Deep SR (MDSR) architecture, however, reduces the number of parameters through a majority of shared parameters. Scale-specific layers are only applied close to the input and output blocks in parallel to learn scale-dependent representations. The proposed deep architectures are trained using $\ell_1$ loss.
Data augmentation (rotations and flips) was used to create a `self-ensemble' \ie, transformed inputs are passed through the network, reverse-transformed and averaged together to create a single output. The authors noted that such a self-ensemble scheme does not require learning multiple separate models, but results in a gain comparable to conventional ensemble based models. EDSR and MDSR achieve better performance, in terms of quantitative measures ( \eg, PSNR), compared to older architectures such as SRCNN, VDSR and other ResNet based closely related architectures (\eg, SRGAN \cite{ledig2017photo}). 

\noindent
$\bullet$ \textbf{CARN:}
Cascading residual network (CARN)~\cite{ahn2018CARN} employs ResNet Blocks~\cite{he2016ResNet} to learn the relationship between 
low-resolution input and high-resolution output. The difference between
the models is the presence of local and global cascading modules.  The features from intermediate layers are cascaded and converged onto a 1$\times$1 convolutional layer.  The local cascading connections are identical to the global cascading connections, except the blocks are simple residual blocks. This strategy makes information propagation efficient due to multi-level representation and many shortcut connections.The architecture of CARN is shown in Figure~\ref{fig:archs}.

The model is trained using 64$\times$64 patches from BSD~\cite{martin2001BSD}, Yang \etal~\cite{yang2010image91} and DIV2K dataset~\cite{timofte2017ntireFlicker2K} with data augmentation, employing $\ell_1$ loss. Adam~\cite{kingma2014adam} is used for optimization with an initial learning rate of 10$^{-4}$ which is halved after every 4 $\times$ 10$^5$ steps.

\subsubsection{Multi-Stage Residual Nets}
A multi-stage design is composed of multiple subnets that are generally trained in succession \cite{jiao2017formresnet,Fan2017BTSRN}. The first subnet usually predicts the coarse features while the other subnets improve the initial predictions. Here, we also include encoder-decoder designs  (\eg, \cite{mao2016REDNet}) that first downsample the input using {an} encoder and then perform upsampling via {a} decoder (hence two distinct stages). The following architectures super-resolved the image in various stages.

\noindent
$\bullet$ \textbf{FormResNet}
is proposed by \cite{jiao2017formresnet} which builds  upon DnCNN as shown in Figure~\ref{fig:archs}. This model is composed of two networks, both of which 
are similar to DnCNN; however, the difference lies in the loss layers. The first network, termed as \enquote{Formatting layer}, incorporates 
Euclidean and perceptual loss. The classical algorithms such as BM3D can also replace this formatting layer. The second deep network \enquote{DiffResNet} is similar to DnCNN and input to this network is fed from the first one. The stated formatting layer removes high-frequency corruption in uniform areas, while DiffResNet learns the structured regions. FormResNet improves upon the results of DnCNN by a small margin. 

\noindent
$\bullet$ \textbf{BTSRN} stands for balanced two-stage residual networks~\cite{Fan2017BTSRN} for image super-resolution. The network is composed of a low-resolution stage and a high-resolution stage.  In the low-resolution stage, the feature maps have a smaller size, the same as the input patch. The feature maps are upsampled using a deconvolution followed by nearest neighbor upsampling. The upsampled feature maps are then fed into the high-resolution stage. In both the low-resolution and the high-resolution stages, a variant of residual block~\cite{he2016deep} called projected convolution is employed. The residual block consists of 1$\times$1 convolutional layer as a feature map projection to decrease the input size of 3$\times$3 convolutional features. The LR stage has six residual blocks while the HR stage consists of four residual blocks.

Being a competitor in the NTIRE 2017 challenge~\cite{timofte2017ntireFlicker2K}, the model is trained on 900 images from DIV2K dataset~\cite{timofte2017ntireFlicker2K}, 800 training image and 100 validation images combined. During training, the images are cropped to 108$\times$108 sized patches and augmented using flipping and rotation operations. The initial learning rate was set to 0.001 which is exponentially decreased after each iteration by a factor of 0.6.  The optimization was performed using Adam~\cite{kingma2014adam}. The residual block consists of 128 feature maps as input and 64 as output. 
$\ell_2$ distance is used for computing difference between the prediction output and the ground-truth.


\noindent
$\bullet$ \textbf{REDNet:}
Recently, due to the success of UNet~\cite{ronneberger2015unet}, 
\cite{mao2016REDNet} proposes {a super-resolution algorithm using} an encoder (based on convolutional layers) and a decoder (based on deconvolutional layers).
REDNet \cite{mao2016REDNet} stands for Residual Encoder Decoder Network and is mainly composed of convolutional and symmetric deconvolutional layers. A rectification layer (ReLU) is added after each convolutional and deconvolutional layer. The convolutional layers extract  feature maps while preserving  object structures and removing  degradations. On the other hand, the deconvolutional layers reconstruct the missing details of the images. Furthermore, skip connections are added between the convolutional and the symmetric deconvolutional layer. The feature maps of the convolutional layer are summed with the output of the mirrored deconvolutional layer before applying non-linear rectification. The input to the network is the bicubic interpolated images, and the outcome of the final deconvolutional layer is a high-resolution image. The proposed network is end-to-end trainable and convergence is achieved by minimizing the $\ell_2$-norm between the output of the system and the ground truth. The architecture of the REDNet \cite{mao2016REDNet} is shown in Figure~\ref{fig:archs}.

The authors proposed three variants of the REDNet architecture where the overall structure remains same, but the number of convolutional and deconvolutional layers are changed. The best performing architecture has 30 weight layers, each with 64 feature maps. Furthermore, the luminance channel from the Berkeley Segmentation Dataset (BSD) \cite{martin2001BSD} is used to generate the training image set. The patches of size 50$\times$50 are extracted with a regular stride as the ground truth, and the input patches are formed from the ground truth by downsampling the patches and then upsampling it to the original size using bicubic interpolation.

The network is trained by extracting patches from 91 images \cite{yang2010image91} and employing Mean square error (MSE) as a loss function.  The input and output patch sizes are 9$\times$9 and 5$\times$5, respectively.  The patches are normalized by its means and variances which are later added to the corresponding restored final high-resolution outputs. Furthermore, the kernel has a size of 5$\times$5 with 
128 feature channels. 

\subsection{Recursive networks}
As the name indicates, recursive networks~\cite{kim2016DRCN,tai2017DRRN,tai2017memnet} either employ recursively connected convolutional layers or recursively linked units.  The main motivation behind these designs is to progressively break down the harder SR problem into a set of simpler ones, that are easy to solve. The basic architecture is shown in Figure~\ref{fig:archs} and we provide further details of recursive models in the following sections.

\subsubsection{DRCN}
As the name indicates, Deep Recursive Convolutional Network (DRCN)~\cite{kim2016DRCN} applies the same convolution layers multiple times.  An advantage of this technique is that the number of parameters remains constant for more recursions. DRCN~\cite{kim2016DRCN} is composed of three smaller networks, termed as embedding net, inference net, and reconstruction net. 

The first sub-net, called the embedding network, converts the input (either grayscale or color image) to feature maps. The subsequent sub-network, known as inference net, performs super-resolution, which analyzes 
image
regions by recursively applying a
single layer consisting of convolution and ReLU. The size of the receptive field is increased after each recursion. The output of the inference net is high-resolution feature maps which are transformed to grayscale or color 
by the reconstruction net.

\subsubsection{DRRN}
Deep Recursive Residual Network (DRRN)~\cite{tai2017DRRN} proposes a deep CNN model but with conservative parametric complexity. Compared to previous models such as VDSR~\cite{kim2016VDSR}, REDNet~\cite{mao2016REDNet} and DRCN~\cite{kim2016DRCN}, this model introduces an even deeper architecture with as many as 52 convolutional layers. At the same time, they reduce the network complexity by 
factors of 14, 6 and 2 for the cases of REDNet, DRCN and VDSR respectively. This is achieved by combining
residual image learning \cite{kim2016accurate} with local identity connections between small blocks of layers with in the network.  The authors stress that such parallel information flow realizes stable training for deeper architectures. 

Similar to DRCN \cite{kim2016DRCN}, DRRN utilizes recursive learning which replicates a basic skip-connection block several times to achieve a multi-path network block (see Figure \ref{fig:archs}).  Since parameters are shared between the replications, the memory cost and computational complexity is significantly reduced. The final architecture is obtained by stacking multiple recursive blocks. DRCN used the standard SGD optimizer with gradient clipping \cite{kim2016accurate} for parameter learning. The loss layer is based on MSE loss, similar to other popular architectures. The proposed architecture reports a consistent improvement over previous methods, which supports the case for deeper recursive architectures and residual learning. 

\subsubsection{MemNet}
\begin{figure*}
    \centering
    \includegraphics[width=\textwidth, clip,trim=0 0.0cm 0.0cm 0]{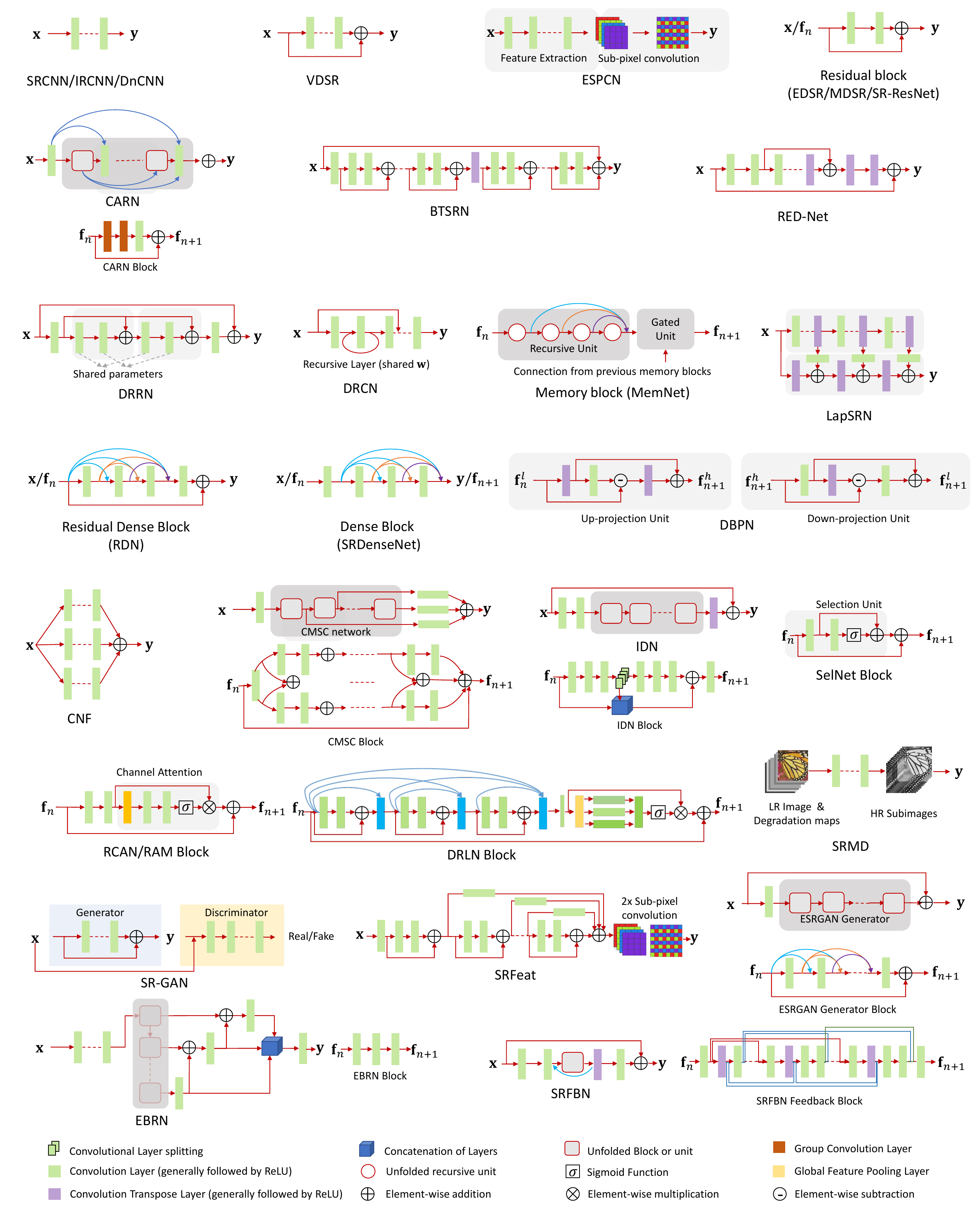}
    \caption{A glimpse of the diverse range of network architectures used for single-image super-resolution using deep networks. The order of the networks is based on their presentation in this paper.}
    \label{fig:archs}
\end{figure*}

A novel persistent memory network for image super-resolution (abbreviated as MemNet) is present by Tai \etal \cite{tai2017memnet}. MemNet can be broken down into three parts similar to SRCNN \cite{dong2014SRCNN}. The \emph{first} part is called the feature extraction block, which extracts features from the input image. This part is consistent with earlier designs such as~\cite{dong2014SRCNN, dong2016FSRCNN, shi2016ESPCN}. The \emph{second} part consists of a series of memory blocks stacked together.   This part plays the most crucial role in this network. The memory block, as shown in Figure~\ref{fig:archs}, consists of a recursive unit and a gate unit. The recursive part is similar to ResNet~\cite{he2016ResNet} and is composed of two convolutional layers with a pre-activation mechanism and dense connections to the gate unit.  Each gate unit is a convolutional layer with 1$\times$1 convolutional kernel size.

The MSE loss function is adopted by MemNet~\cite{tai2017memnet}.  The experimental settings are the same as VDSR \cite{kim2016VDSR}, using 200 images from BSD~\cite{martin2001BSD} and 91 images from Yang \etal \cite{yang2010image91}. The network consists of six memory blocks with six recursions. The total number of layers in 
MemNet is 80.  MemNet is also employed for other image restoration tasks such as image denoising, and JPEG deblocking where it shows promising results. 

\subsubsection{SRFBN} Li~\etal~\cite{li2019feedback} proposed a Super-resolution Feedback Network (SRFBN) based on a recurrent architecture design. Specifically, the low-resolution input is recursively refined to obtain a corresponding high-resolution output. The main architecture is based on a feedback block (FB) that consists of several projection groups. Each projection group first finds high-resolution features (via deconvolution) and then generates low-resolution features (via convolution). Their exist dense connections between the low-resolution and high-resolution representations within each FB. At different time-steps, inputs are recursively passed to the FB, which learns the residual signal due to the existence of a global residual connection.

SRFBN is trained using a curriculum learning approach for the case when multiple types of degradations exist in the LR image. In this process, HR images with increasing complexity are presented to the model as ground-truth. The model is trained with a $\ell_1$ objective, and a total of four recursive iterations are used during training. The evaluations are reported for other degradations (e.g., Gaussian blur) in addition to usual bicubic downsampling. The recursive design allows this approach to work with a relatively less number of trainable parameters.

\subsection{Progressive reconstruction designs}
Typically, 
CNN algorithms predict the output in one step; however, it may not be feasible for large scaling factors. To deal with large factors, some algorithms~\cite{wang2015SCN,lai2017LapSRN}, predict the output in multiple steps \ie 2$\times$ followed by 4$\times$ and so on. Here, we introduce such algorithms.

\subsubsection{SCN}
Wang \etal \cite{wang2015SCN} proposed a scheme which consolidates the merits of sparse coding \cite{yang2010Sparse} with domain knowledge of deep neural networks. With this combination, it aims for a compact model and improved performance. The proposed sparse coding-based network (SCN) \cite{wang2015SCN} mimics a Learned Iterative Shrinkage and Thresholding Algorithm (LISTA) network to build a multi-layer neural network.

Similar to SRCNN \cite{dong2016SRCNNPAMI}, the first convolutional layer extracts features from the low-resolution patches which are then fed into a LISTA network. To obtain the sparse code for each feature, the LISTA network consists of a finite number of recurrent stages.  The LISTA stage is composed of two linear layers and a nonlinear layer with an activation function having a threshold which is learned/updated during training. To simplify training, the authors decomposed the nonlinear neuron into two linear scaling layers and a unit-threshold neuron. The two scaling layers are diagonal matrices which are reciprocal to each other \eg if multiplication scaling layer is present, division after the threshold unit follows it. After the LISTA network, the original high-resolution patches are reconstructed by multiplying the sparse code and high-resolution dictionary in the successive linear layer. As a final step, again using a linear layer, the high-resolution patches are placed in the original location in the image to obtain the high-resolution output. 

\subsubsection{LapSRN}
Deep Laplacian pyramid super-resolution network (LapSRN) \cite{lai2017LapSRN} employs a pyramidal framework. LapSRN consists of three sub-networks that progressively predict the residual images up to a factor of 8$\times$. The residual images of each sub-network are added to the input LR image to obtain 
SR images. The output of the first sub-network is a residue of 2$\times$, the second sub-network provides a 4$\times$ residue, and the last one gives the 8$\times$ residual image.  These residual images are added to the correspondingly scaled upsampled images to obtain the final super-resolved images. The authors term the residual prediction branch as feature extraction while the addition of bicubic images with the residue is called image reconstruction branch. The Figure \ref{fig:archs} shows the LapSRN network which consists of three types of elements \ie the convolutional layers, leaky ReLU, and deconvolutional layers.  Following the CNN convention, the convolutional layers precede the leaky ReLU (allowing a negative slope of 0.2) and deconvolutional layer at the end of the sub-network to increase the size of the residual image to the corresponding scale.

LapSRN uses a differentiable variant of $\ell_1$ loss function known as Charbonnier which can handle outliers. The loss is employed at every sub-network, resembling a multi-loss structure. Furthermore, the filter sizes for convolutional and deconvolutional layers are 3$\times$3 and 4$\times$4, respectively, having 64 channels each. The training data is similar to SRCNN \cite{dong2014SRCNN} \ie 91 images from  Yang~\etal~\cite{yang2010image91} and 200 images from BSD dataset \cite{martin2001BSD}.

The LapSRN model uses three distinct models to perform 2$\times$, 4$\times$ and 8$\times$ SR. They also propose a single model, termed as Multi-scale (MS) LapSRN, that jointly learns to handle multiple SR scales \cite{MSLapSRN}. Interestingly, a single MS-LapSRN model outperforms the results obtained from three distinct models. One explanation for this effect is that the single model leverages common inter-scale traits that help in achieving more accurate results. 

\subsection{Densely Connected Networks}
Inspired by the success of the DenseNet~\cite{huang2017densely} architecture for image classification, super-resolution algorithms based on densely connected CNN layers have been proposed to improve 
performance. The main motivation in such a design is to combine hierarchical cues available along the network depth to achieve high flexibility and richer feature representations. We discuss some popular designs in this category below. 

\subsubsection{SRDenseNet}
This network architecture \cite{tong2017image} is based on the DenseNet \cite{huang2017densely} which uses dense connections between the layers \ie a layer directly operates on the output from all previous layers. Such an information flow from low to high-level feature layers avoids the vanishing gradient problem, enables learning compact models and speeds up the training process. Towards the rear part of the network, SRDenseNet uses a couple of deconvolution layers to upscale the inputs.  The authors propose three variants of 
SRDenseNet, (1) a sequential arrangement of dense blocks followed by deconvolution layers. In this way only high-level features are used for reconstructing the final SR image. (2) Low-level features from initial layers are combined before final reconstruction. For this purpose, a skip connection is used to combine low- and high-level features. (3) All features are combined by using multiple skip connections between low-level features and the dense blocks to allow a direct flow of information for a better HR reconstruction. Since complementary features are encoded at multiple stages in the network, the combination of all feature maps gives the best performance among other variants of SRDenseNet. The MSE error ($\ell_2$ loss) is used as a loss to train the full model. Overall, SRDenseNet models demonstrate a consistent improvement in performance over the models that do not use dense connections between layers.  

\subsubsection{RDN}
As the name implies, Residual Dense Network  \cite{zhang2018RDN} (RDN) combines residual skip connections (inspired by SRResNet) with dense connections (inspired by SRDenseNet). The main motivation is that the hierarchical feature representations should be fully used to learn local patterns. To this end, residual connections are introduced at two levels; local and global. At the local level, a novel residual dense block (RDB) was proposed where the input to each block (an image or output from a previous block) is forwarded to all layers with in the RDB and also added to the block's output so that each block focuses more on the residual patterns. Since the dense connections quickly lead to high dimensional outputs, a local feature fusion approach to reduce the dimensions with $1{\times}1$ convolutions was used in each RDB. At the global level, outputs of multiple RDBs are fused together (via concatenation and $1{\times}1$ convolution operations) and a global residual learning is performed to combine features from multiple blocks in the network. The residual connections help stabilize 
network training and results in an improvement over the SRDenseNet \cite{tong2017image}. 

In contrast to the $\ell_2$ loss used in SRDenseNet, RDN utilizes the $\ell_1$ loss function and advocates its improved convergence properties.  
Network training is performed on $32{\times}32$ patches randomly selected in each batch. Data augmentation by flips and rotations is applied as a regularization measure. The authors also experiment with settings where different forms of degradation (\eg., noise and artifacts) are present in LR images. The proposed approach shows good resilience against such degradation and recovers much enhanced SR images.

\subsubsection{D-DBPN}
Dense deep back-projection network for super-resolution \cite{haris2018DDBPN} takes inspiration from the conventional SR approaches (\eg, \cite{irani1991improving}) that iteratively perform back-projections to learn the feedback error signal between LR and HR images. The motivation is that only a feed-forward approach is not optimal for modelling the mapping from LR to HR images, and a feedback mechanism can greatly help in achieving better results. For this purpose, the proposed architecture comprises of a series of up and down sampling layers that are densely connected with each other. In this manner, HR images from multiple depths in the network are combined to achieve the final output.

The architecture of up and down sampling blocks is shown in Fig.~\ref{fig:archs}. For the sake of brevity, the simpler case of single connection from previous layers is shown, and the readers are directed to \cite{haris2018DDBPN} for the complete densely connected block. An important feature of this design is the combination of upsampling outputs for input feature map and the residual signal. The explicit addition of residual signal in the upsampled feature map provides error feedback and forces the network to focus on fine details. The network is trained using the standard $\ell_1$ loss function. D-DBPN has a relatively high computational complexity of $\sim 10$ million parameters for 4$\times$ SR, however a lower complexity version of the final model was also proposed that led to a slight drop in performance.

\subsection{Multi-branch designs}
In contrast to single-stream (linear) and skip-connection based designs, multi-branch networks aim to obtain a diverse set of features at multiple context scales. Such complementary information is then fused to obtain better HR reconstructions. This design also enables a multi-path signal flow, leading to better information exchange in forward-backward steps during training. Multi-branch designs are becoming common in several other computer vision tasks as well. 
We explain multi-branch networks in the section below.

\subsubsection{CNF}
Ren \etal~\cite{Ren2017CNF} proposed fusing multiple convolutional neural networks for image super-resolution. The authors termed their CNN network Context-wise Network Fusion (CNF), where each SRCNN~\cite{dong2014SRCNN} is constructed with a different number of layers. The output of each SRCNN~\cite{dong2014SRCNN} is then passed through a single convolutional layer and eventually all of them are fused using sum-pooling. 

The model is trained on 20 million patches collected from Open Image Dataset~\cite{Ivan2017OpenImages,Kuznetsova2018OpenImages}. The size of each patch is 33$\times$33 pixels of luminance channel only.  First, each SRCNN is trained individually for 50 epochs with a learning rate of 1e-4; then the fused network is trained for ten epochs with the same learning rate. Such a progressive learning strategy is similar to curriculum learning that starts from a simple task and then moves on to the more complex task of jointly optimizing multiple sub-nets to achieve improved SR. Mean square error is used as a loss for the network training.

\subsubsection{CMSC}
Cascaded multi-scale cross-network, abbreviated as CMSC \cite{hu2018CMSC}, is composed of a feature extraction layer, cascaded subnets, and a reconstruction network. The feature extraction layer performs the same function as mentioned for the cases of SRCNN \cite{dong2014SRCNN}, FSRCNN \cite{dong2016FSRCNN}. Each subnet is composed of merge-and-run (MR) blocks.   Each MR block is comprised of two parallel branches having two convolutional layers each. The residual connections from each branch are accumulated together and then added to the output of both branches individually as shown in Figure \ref{fig:archs}. Each subnet of CMSC is formed with four MR blocks having different receptives field of 3$\times$3, 5$\times$5, and 7$\times$7 to capture contextual information at multiple scales.  Furthermore, each convolutional layer in the MR block is followed by batch normalization and Leaky-ReLU \cite{maas2013LeakyReLU}. The last reconstruction layer generates the final output.

The loss function is $\ell_1$ which combines the intermediate outputs with the final one using a balancing term. The input to the network is upsampled using bicubic interpolation with a patch size of 41 $\times$ 41. The model is trained with 291 images similar to VDSR \cite{kim2016VDSR} using an initial learning rate of 10$^{-1}$, decreasing by a factor of 10 after every ten epochs for a total of 50 epochs. CMSC lags in performance compared to EDSR \cite{lim2017EDSR} and its variant MDSR \cite{lim2017EDSR}.

\subsubsection{IDN}
The Information Distillation Network (IDN)~\cite{hui2018fast} consists of three blocks: a \emph{feature extraction} block, multiple stacked \emph{information distillation} blocks and a \emph{reconstruction} block. The feature extraction block is composed of two convolutional layers to extract features. The distillation block is made up of two other blocks, an enhancement unit, and a compression unit.  The enhancement unit has six convolutional layers followed by leaky ReLU. The output of the third convolutional layer is sliced, the half batch is concatenated with the input of the block, and the other half is used as an input to the fourth convolutional layer. The output of the concatenated component is added with the output of the enhancement block. In total, four enhancement blocks are utilized. The compression unit is realized using a 1$\times$1 convolutional layer after each enhancement block. The reconstruction block is a deconvolution layer with a kernel size of 17$\times$17.

The network is first trained using absolute mean error loss and then fine-tuned by the mean square error loss. The images of training are the same as~\cite{tai2017memnet}. The input patch size is 26 $\times$ 26.  The initial learning rate is set to be 1e-4 for a total of 10$^5$ iterations, utilizing Adam~\cite{kingma2014adam} as an optimizer.

\subsubsection{EBRN} The Embedded Block Residual Network (EBRN) \cite{qiu2019embedded} is based on the idea that different frequencies occurring in an image require different levels of processing. For example, low-frequency information can be restored by a shallow network, while more complex, high-frequency content would need a deeper network for accurate modeling. To this end, they propose a multi-branch architecture where the more-complex signal is passed onto deeper modules during super-resolution.

The final model comprises of ten Block Residual Modules (BRM), resulting in a total of ten parallel branches in the model. The new fusion strategy is used to combine outputs from multiple branches. Instead of simple summation of all outputs, a recursive fusion approach is used where only the outputs from two neighboring branches are progressively combined until they reach the main low-resolution branch.  The model is first trained with a $\ell_1$ loss and then fine-tuned with a $\ell_2$ objective to penalize the outliers in predicted outputs.   Overall, the proposed approach delivers impressive results compared to the state of the art models on $2\times$, $4\times$, and $8\times$ super-resolution.

\subsection{Attention-based Networks}
The previously discussed network designs consider all spatial locations and channels to have a uniform importance for the super-resolution. In several cases, it helps to selectively attend to only a few features at a given layer. Attention-based models~\cite{Choi2017SelNet,zhang2018RCAN} allow this flexibility and consider that not all the features are essential for super-resolution but have varying importance. Coupled with deep networks, recent attention-based models have shown significant improvements for SR. Following are the examples of CNN algorithms using attention mechanisms.

\subsubsection{SelNet}
Choi and Kim~\cite{Choi2017SelNet} proposed a novel selection unit for the image super-resolution network, termed as SelNet. The selection unit serves as a gate between convolutional layers, allowing only selected values from the feature maps. The selection unit is composed of an identity mapping and a cascade of ReLU, 1$\times$1 convolution and a sigmoid layer. SelNet consists of a total of 22 convolutional layers, and the selection unit is added after every convolutional layer. Similar to VDSR~\cite{kim2016VDSR}, residual learning and gradient switching (a version of gradient clipping) are also employed in SelNet~\cite{Choi2017SelNet} for faster learning.

The low-resolution patches of size 120$\times$120 are input to the network which are cropped from DIV2K dataset~\cite{timofte2017ntireFlicker2K}. The number of epochs is set to 50 with a learning rate of 10$^{-1}$. The loss used for training the SelNet is $\ell_2$. 

\subsubsection{RCAN}
Residual Channel Attention Network (RCAN)~\cite{zhang2018RCAN} is a recently proposed deep CNN architecture for single image super-resolution. The main highlights of the architecture include: (a) a recursive residual design where residual connections exist within each block of a global residual network and (b) each local residual block has a channel attention mechanism such that the filter activations are collapsed from $h\times w \times c$ to a vector with $1 \times 1 \times c$ dimensions (after passing through a bottleneck) that acts as a selective attention over channel maps. The first novelty allows multiple pathways for information flow from initial to final layers. The second contribution allows the network to focus on selective feature maps that are more important for the end task and also effectively models the relationships between feature maps. 

RCAN~\cite{zhang2018RCAN} uses $\ell_1$ loss function for network training. It was observed that the recursive residual style architecture leads to better convergence properties of very deep networks. Furthermore, it leads the better performance compared to contemporary approaches such as IRCNN~\cite{zhang2017IrCNN}, VDSR~\cite{kim2016VDSR} and RDN~\cite{zhang2018RDN}. This shows the effectiveness of channel attention mechanisms~\cite{hu2017squeeze} for low-level vision tasks. Having said that, one shortcoming of the proposed framework is its high computational complexity ($\sim15$ million parameters for 4$\times$ SR) compared to \eg LapSRN~\cite{lai2017LapSRN}, MemNet~\cite{tai2017memnet} and VDSR~\cite{kim2016VDSR}.

\subsubsection{DRLN}
More recently, densely residual Laplacian attention Network (DRLN)~\cite{anwar2019DRLN} is introduced to super-resolve the images.  The network structure is modular and hierarchal, and the main highlights of the network are 1): modular architecture, 2): densely connected residual units,  3): Cascading connections,  and 4): Laplacian attention. DRLN~\cite{anwar2019DRLN} exploits difference connections such as long-skips, medium-skips, local-skips alongside the cascaded ones. Similarly, in each block, three residual units are densely connected to learn a compact representation.  Then, the learned features are weighted using Laplacian attention in the same block. The structure is repeated throughout the network in each block. Currently, the best results for all datasets are provided by DRLN.

Similar to RCAN~\cite{zhang2018RCAN}, DRLN~\cite{anwar2019DRLN} adopts $\ell_1$ loss function to train the network. The settings for training are the same as RCAN~\cite{zhang2018RCAN} \ie the training patch size, the number of epochs, optimizer \etc The improvement of DRLN~\cite{anwar2019DRLN} can be attributed to the innovative module with Laplacian attention and cascading structure. The number of convolutional layers of DRLN~\cite{anwar2019DRLN} is significantly less as compared to the RCAN. While, on the other hand, the number of parameters of DRLN~\cite{anwar2019DRLN} is higher; however, it is computationally inexpensive due to concatenation of the channels in contrast to RCAN~\cite{zhang2018RCAN} where expensive operation \ie channel addition is used.  

\subsubsection{SRRAM}
This recent work \cite{kim2018ram} focuses on the attention blocks used for
single image super-resolution. They evaluate a range of attention mechanisms with common SR architectures to compare their performance and individual merits/demerits. A Residual Attention Module for SR (SRRAM) is proposed. The structure of SRRAM~\cite{kim2018ram} is similar to RCAN~\cite{zhang2018RCAN}, as both these methods are inspired from EDSR~\cite{lim2017EDSR}. The SRRAM can be divided into three parts which are \emph{feature extraction}, \emph{feature upscaling} and \emph{feature reconstruction}. The first and the last part are similar to the previously discussed methods~\cite{dong2016SRCNNPAMI,dong2016FSRCNN}. However, the feature upscaling part is composed of residual attention modules (RAM). The RAM is a basic unit of SRRAM which is formed of residual blocks followed by spatial attention and channel attention for learning the inter-channel and intra-channel dependencies.

The model is trained using randomly cropped 48$\times$48 patches from DIV2K dataset~\cite{timofte2017ntireFlicker2K} with data augmentation. The filters are of 3$\times$3 size with feature maps of 64. The optimizer used is Adam~\cite{kingma2014adam} employing $\ell_1$ loss, fixing the initial learning rate as 10$^{-4}$. There are a total of 64 RAM blocks used in the final model.

\begin{figure*}
\begin{center}
\begin{tabular}{c@{}c@{}c@{}c@{} c@{} c@{}c@{}c@{}c}
     \raisebox{2\normalbaselineskip}[0pt][0pt]{\rotatebox{90}{Set5}}&
    \includegraphics[width=.13\textwidth, height=.1\textheight]{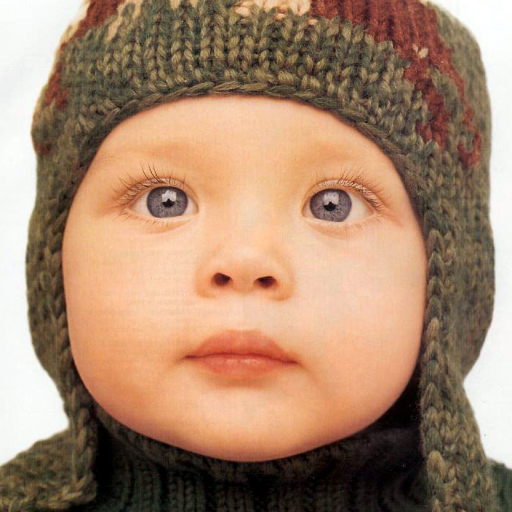}&  
    \includegraphics[width=.13\textwidth, height=.1\textheight]{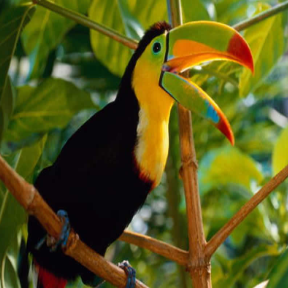}&  
    \includegraphics[width=.13\textwidth,  height=.1\textheight]{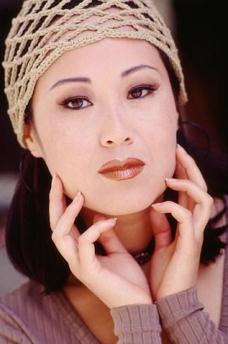}&
    ~~~~~~&
    \raisebox{2\normalbaselineskip}[0pt][0pt]{\rotatebox{90}{Set14}}&
    \includegraphics[width=.13\textwidth, height=.1\textheight]{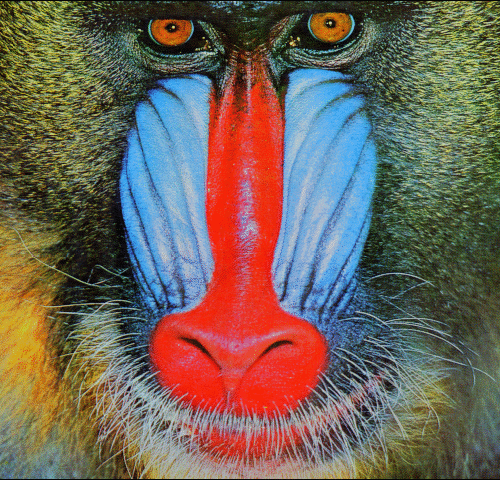}&  
    \includegraphics[width=.13\textwidth, height=.1\textheight]{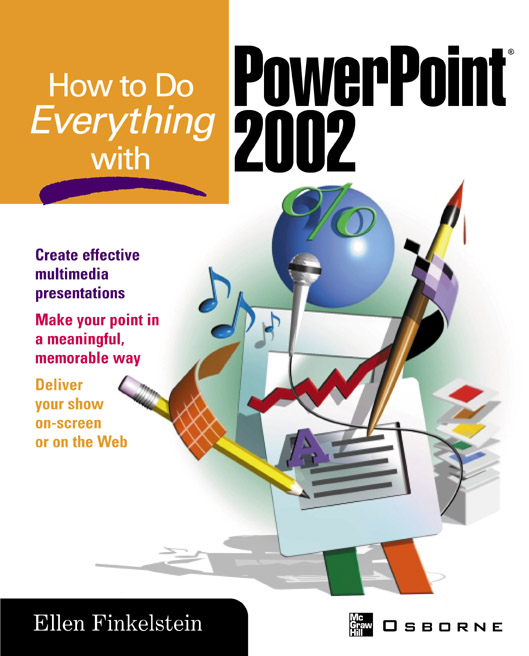}&  
    \includegraphics[width=.13\textwidth, height=.1\textheight]{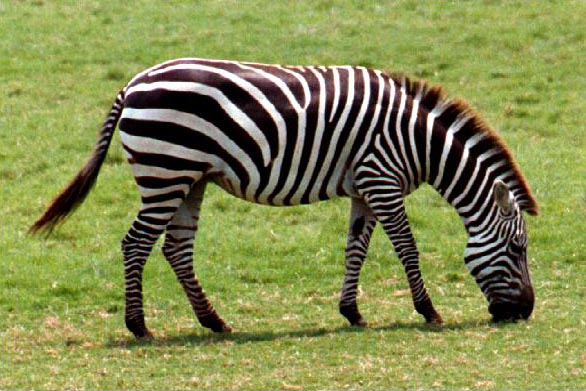}\\  
    
    \raisebox{1.5\normalbaselineskip}[0pt][0pt]{\rotatebox{90}{Urban100}}&
    \includegraphics[width=.13\textwidth, height=.1\textheight]{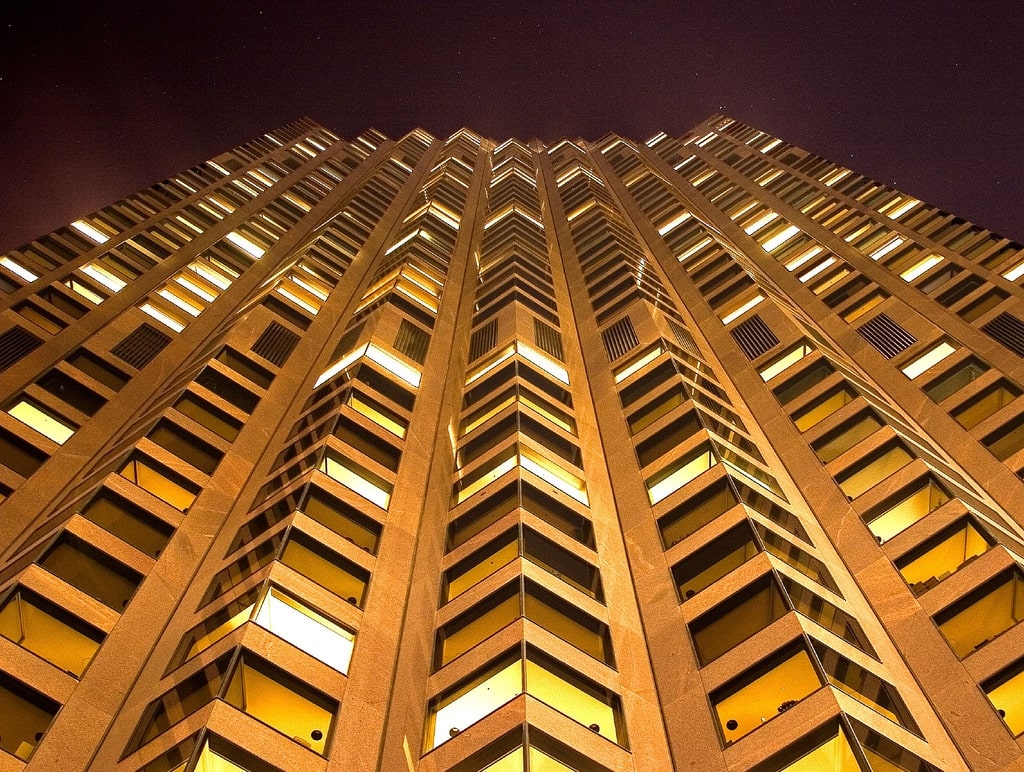}&
    \includegraphics[width=.13\textwidth, height=.1\textheight]{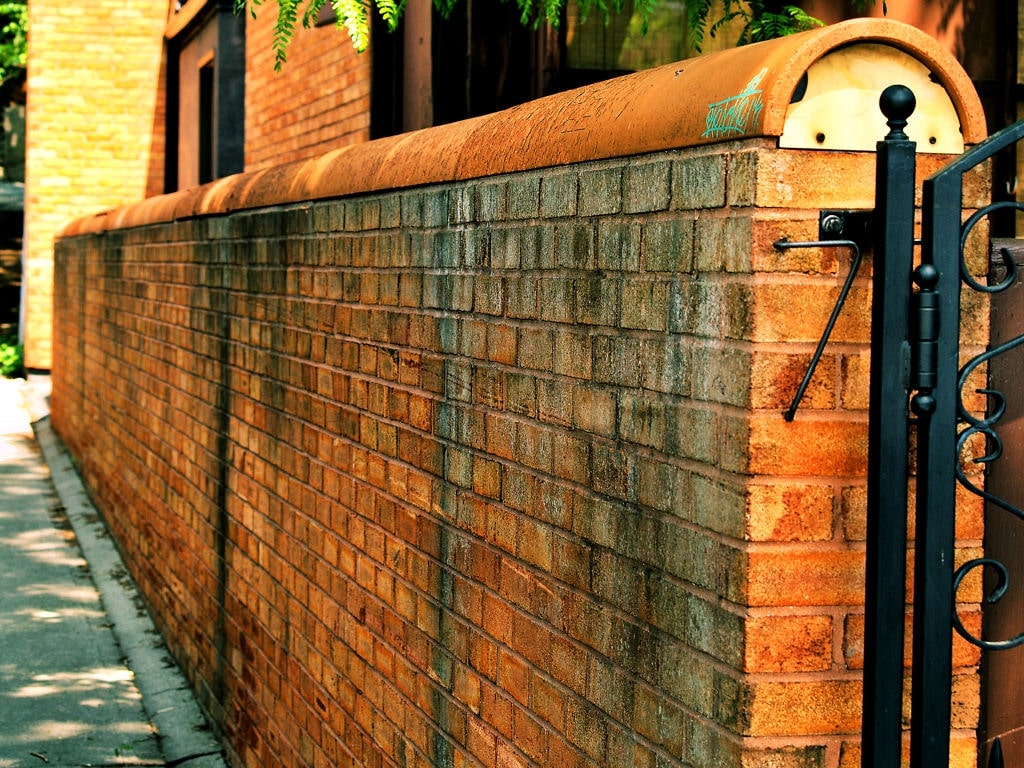}& 
    \includegraphics[width=.13\textwidth, height=.1\textheight]{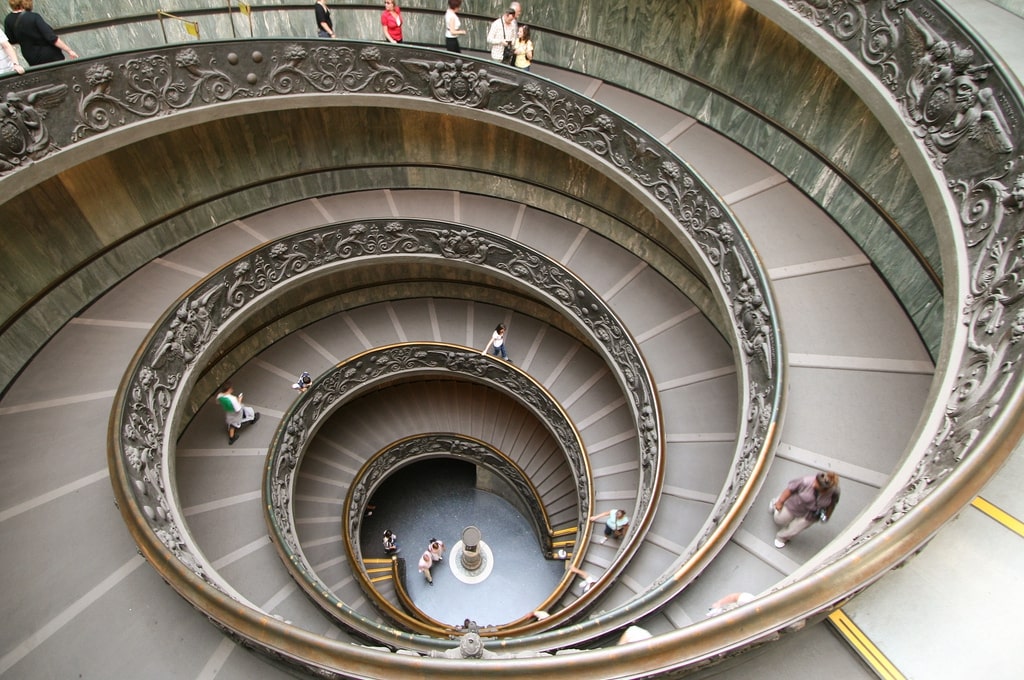}&
    
    ~~~~~~&
    \raisebox{1.5\normalbaselineskip}[0pt][0pt]{\rotatebox{90}{BSD100}}&
    \includegraphics[width=.13\textwidth, height=.1\textheight]{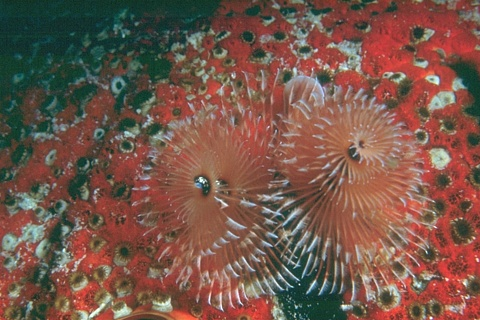}&  
    \includegraphics[width=.13\textwidth, height=.1\textheight]{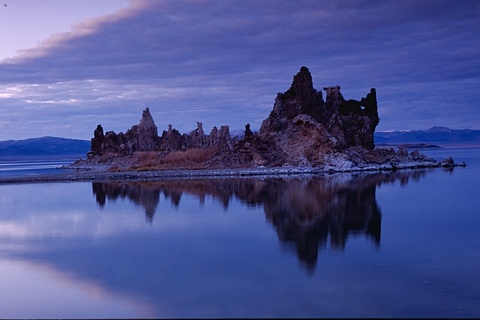}&  
    \includegraphics[width=.13\textwidth, height=.1\textheight]{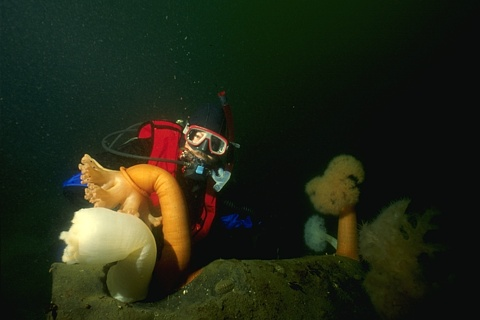}\\
    
    \raisebox{1.5\normalbaselineskip}[0pt][0pt]{\rotatebox{90}{DIV2K}}&
    \includegraphics[width=.13\textwidth, height=.1\textheight]{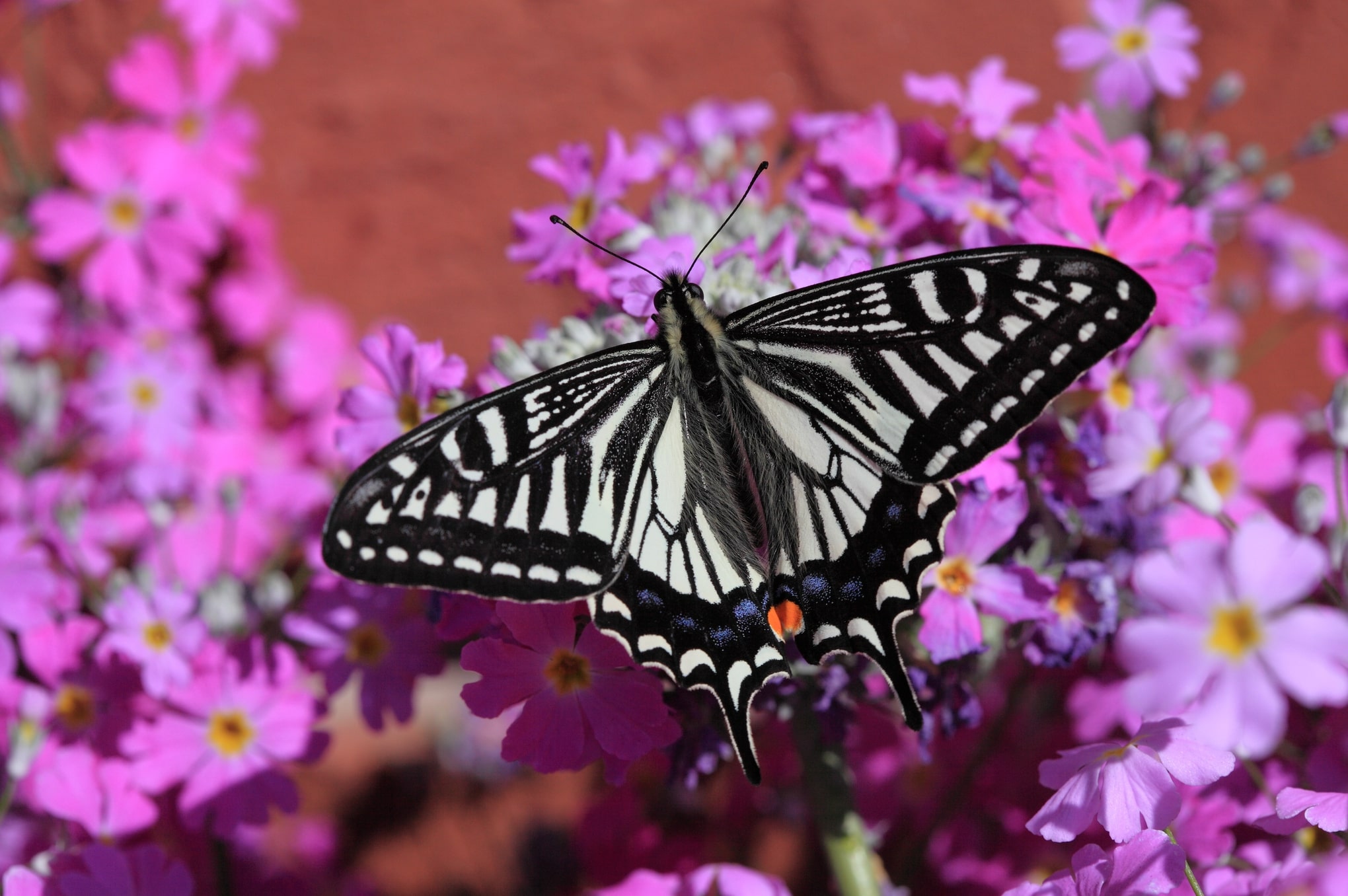}&  
    \includegraphics[width=.13\textwidth, height=.1\textheight]{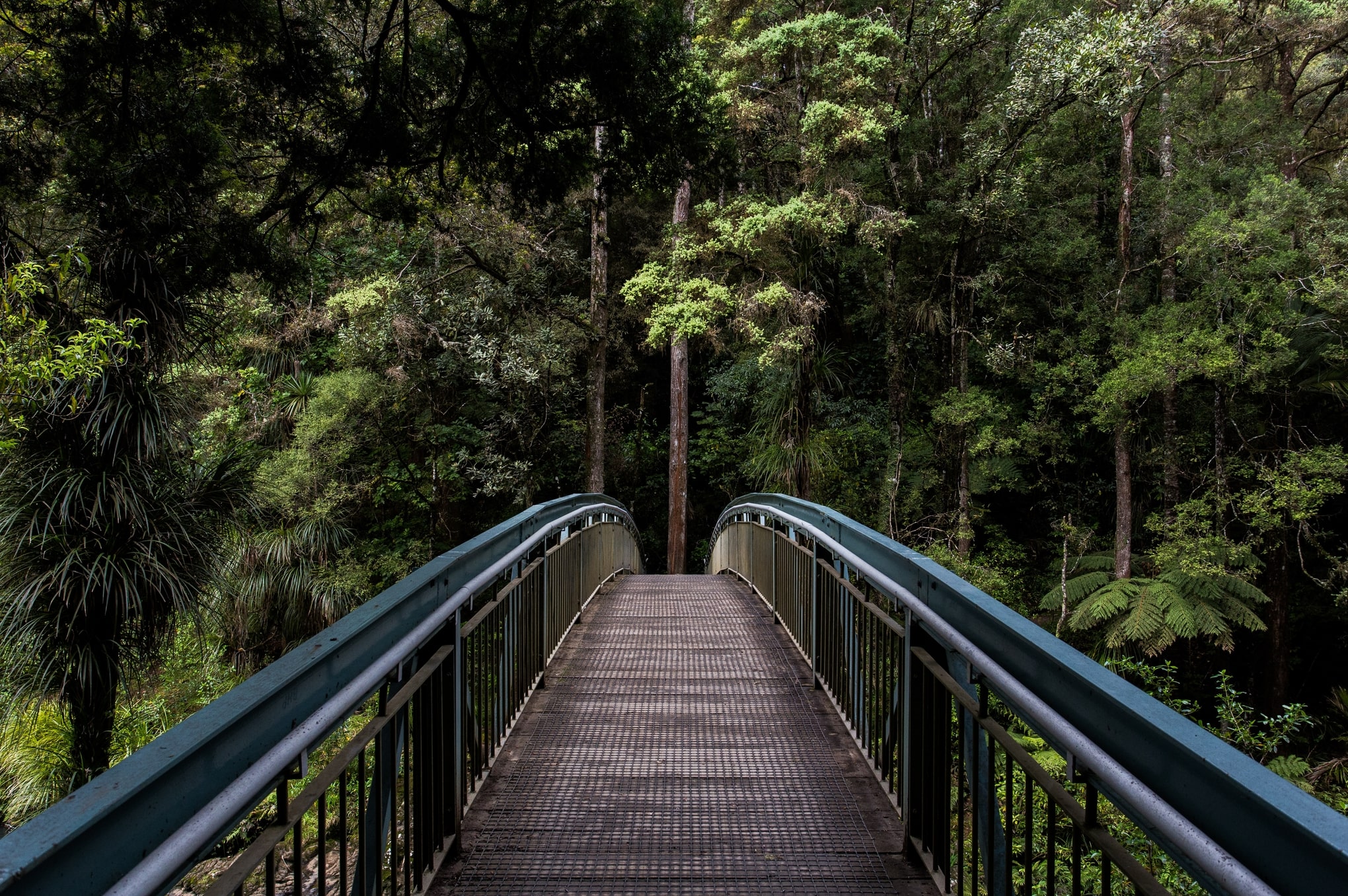}&  
    \includegraphics[width=.13\textwidth, height=.1\textheight]{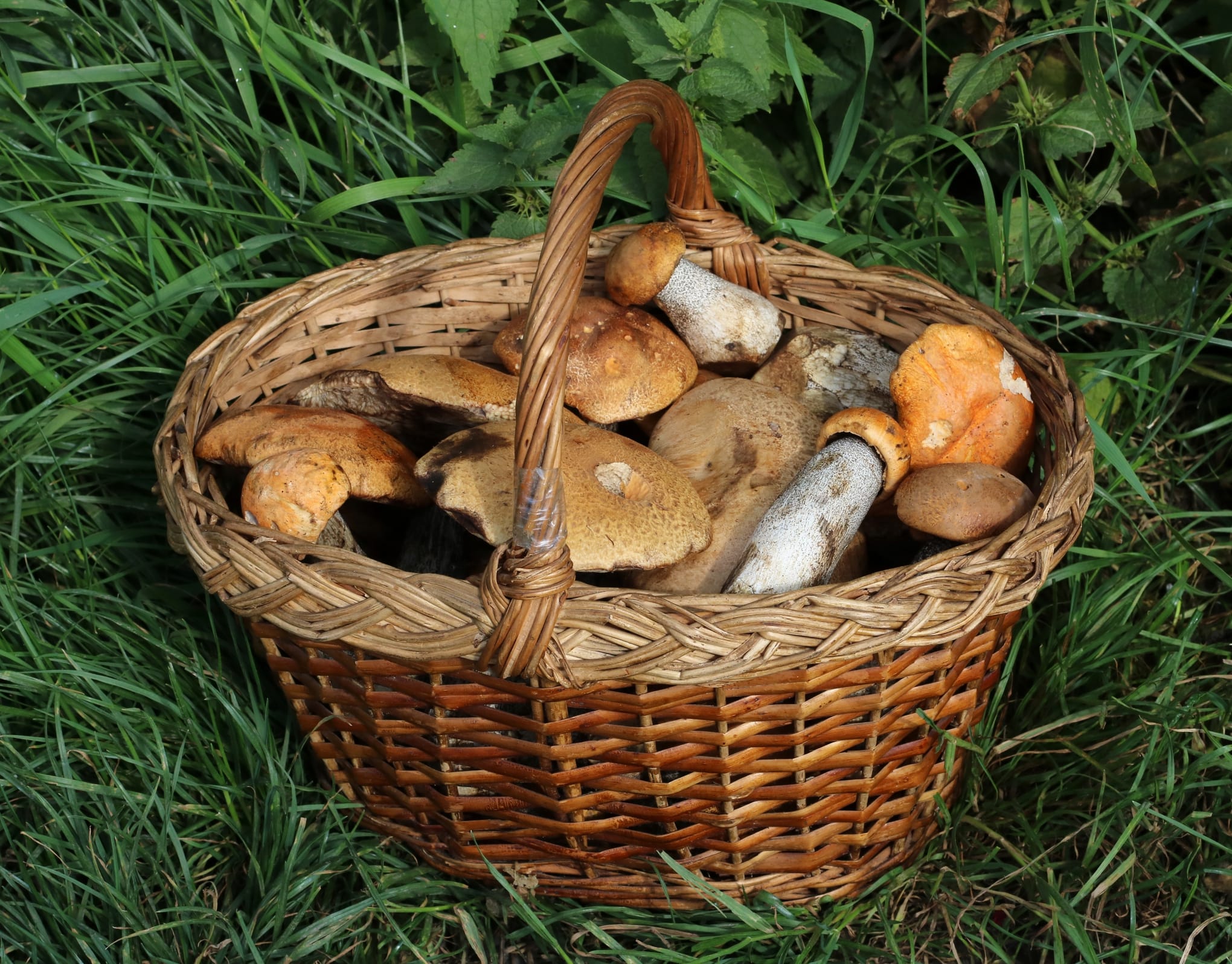}&
    
    ~~~~~~&
    \raisebox{1\normalbaselineskip}[0pt][0pt]{\rotatebox{90}{Manga109}}&
    \includegraphics[trim={30cm 0 0 0},clip,width=.13\textwidth, height=.1\textheight]{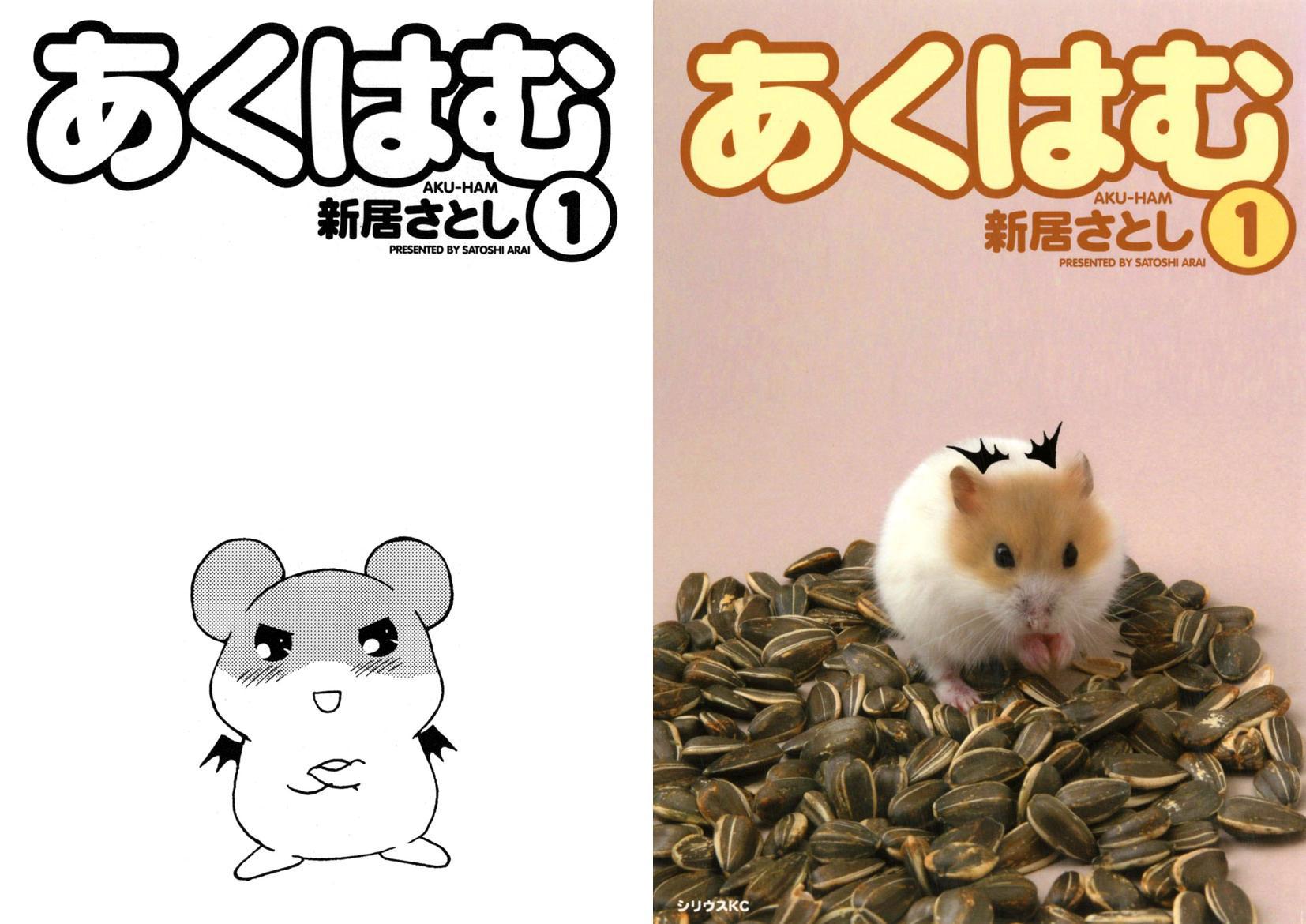}&  
    \includegraphics[trim={30cm 0 0 0},clip,width=.13\textwidth, height=.1\textheight]{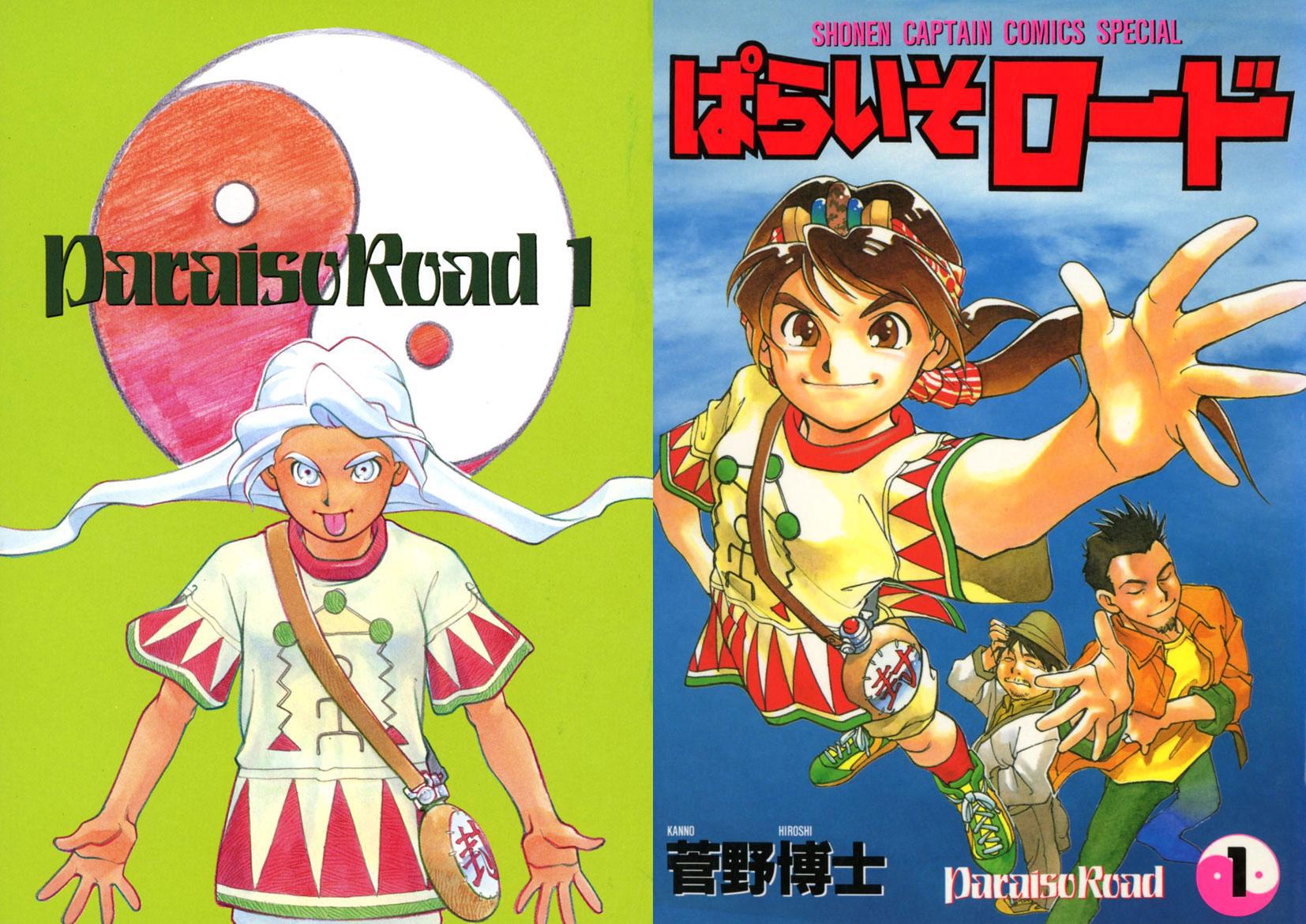}&  
    \includegraphics[trim={30cm 0 0 0},clip,width=.13\textwidth, height=.1\textheight]{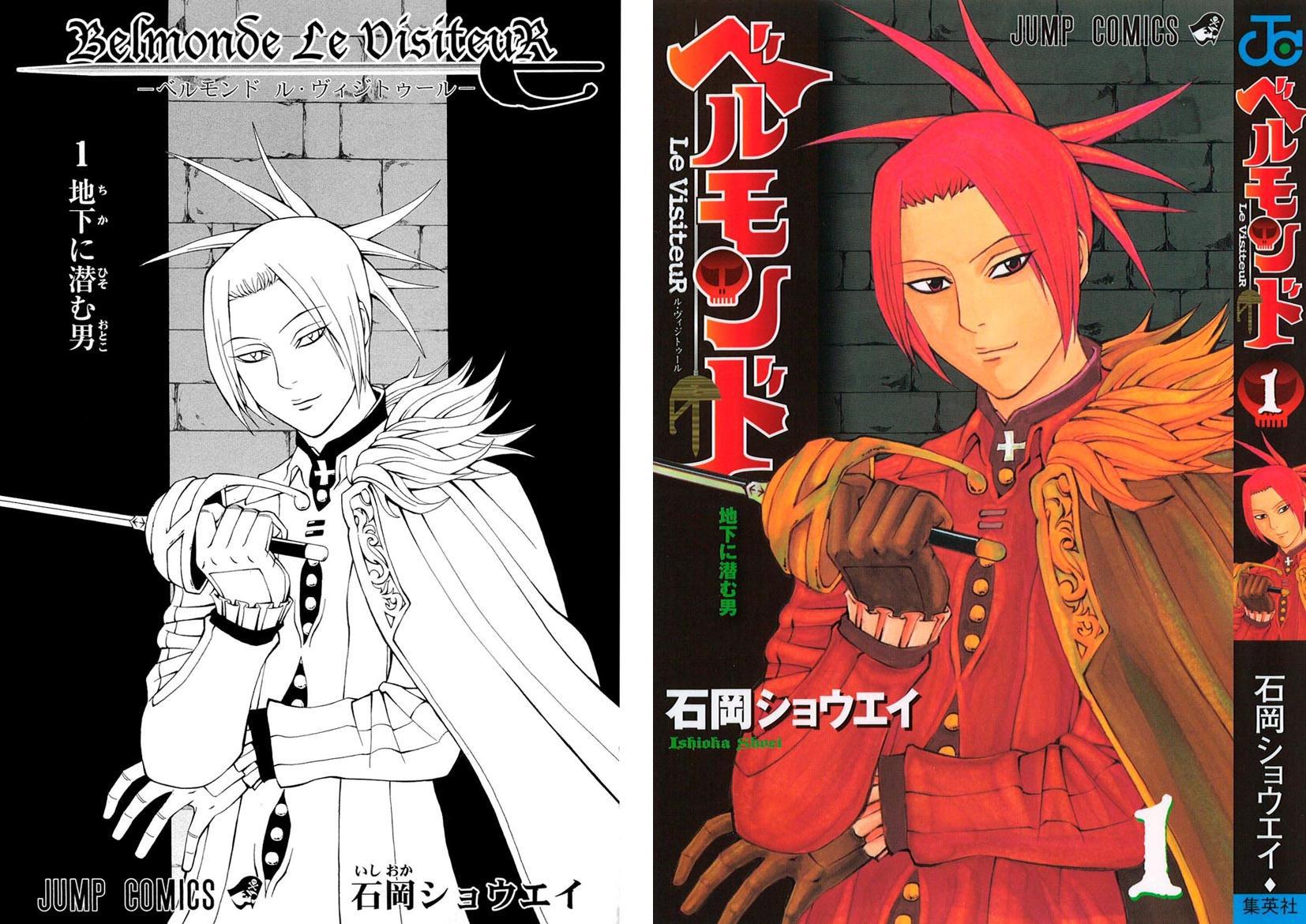}\\ 
    
 \end{tabular}
\end{center}
\caption{Representative test images from six super-resolution datasets used for comparing and evaluating algorithms.}
\label{fig:dataset_images}
\end{figure*}

\subsection{Multiple-degradation handling networks}
The super-resolution networks discussed so far (\eg,~\cite{dong2016SRCNNPAMI,kim2016VDSR}) consider bicubic degradations. However, in reality, this may not be a feasible assumption as multiple degradations can simultaneously occur. To deal with such real-world scenarios, the following methods are proposed.

\subsubsection{ZSSR}
stands for Zero-Shot Super-Resolution \cite{shocher2017ZSSR} and it follows the footsteps of classical methods by super-resolving the images using the internal image statistics employing the power of deep neural networks. The ZSSR \cite{shocher2017ZSSR} uses a simple network architecture that is trained using a downsampled version of the test image. The aim here is to predict the test image from the LR image created from the test image. Once the network learns the relationship between the LR test image and the test image, the same network is used to predict the SR image using the test image as an input.  Hence it does not require training images for a particular degradation and can learn an image-specific network on-the-fly during inference. The ZSSR \cite{shocher2017ZSSR} has a total of eight convolutional layers followed by ReLU consisting of 64 channels. Similar to \cite{kim2016VDSR,lim2017EDSR}, ZSSR \cite{shocher2017ZSSR} learns the residue image using $\ell_1$ norm.

\subsubsection{SRMD}
Super-resolution network for multiple degradations (SRMD) \cite{zhang2018SRMDNF} takes a concatenated low-resolution image and its degradation maps. The architecture of SRMD is similar to \cite{zhang2017IrCNN,zhang2017DnCNN,dong2016SRCNNPAMI}. First, a cascade of convolutional layers of 3$\times$3 filter size is applied to extracted features, followed by a sequence of Conv, ReLU and Batch normalization layers. Furthermore, similar to \cite{shi2016ESPCN}, a convolution operation is utilized to extract HR sub-images, and as a final step, the multiple HR sub-images are transformed to the final single HR output. SRMD directly learns HR images instead of the residue of the images. The authors also introduced a variant called SRMDNF, which learns from noise-free degradations. In SRMDNF network, the connections from the first noise-level maps in the convolutional layers are removed; however, the rest of the architecture is similar to SRMD. The network architecture of the SRMD is presented in Figure~\ref{fig:archs}.

The authors trained individual models for each upsampling scale in contrast to the multi-scale training. $\ell_1$ loss is employed, and the size of the training patches is set to 40$\times$40. The number of convolution layers is fixed to 12, while each layer has 128 feature maps. Training is performed on 5,944 images from BSD \cite{martin2001BSD}, DIV2K \cite{timofte2017ntireFlicker2K} and Waterloo \cite{ma2017waterloo} datasets. The initial learning is fixed at $10^{-3}$ which is later decreased to $10^{-5}$. The criteria for learning rate reduction is based on the error change between successive epochs. Both SRMD and its variant are unable to break the PSNR record of earlier SR networks such as EDSR \cite{lim2017EDSR}, MDSR \cite{lim2017EDSR} and CMSC \cite{hu2018CMSC}. However, its ability to jointly tackle multiple degradations offer a unique capability.

\subsection{GAN Models}
Generative Adversarial Networks (GAN) \cite{goodfellow2014generative,radford2015supervisedGAN} employ a game-theoretic approach where two components of the model, namely a generator and discriminator, try to fool the later. The generator creates SR images that a discriminator cannot distinguish as a real HR image or an artificially super-resolved output. In this manner, HR images with better perceptual quality are generated. The corresponding PSNR values are generally degraded, which highlights the problem that prevalent quantitative measures in SR literature do not encapsulate perceptual soundness of generated HR outputs. 
 The super-resolution methods~\cite{ledig2017photo,sajjadi2017enhancenet} based on the GAN framework are explained next.

\subsubsection{SRGAN}
Single image super-resolution by large up-scaling factors is very challenging. SRGAN \cite{ledig2017photo} proposed to use an adversarial objective function that promotes super-resolved outputs that lie close to the manifold of natural images. The main highlight of their work is a multi-task loss formulation that consists of three main parts: (1) a MSE loss that encodes pixel-wise similarity, (2) a perceptual similarity metric in terms of a distance metric defined over high-level image representation (\eg, deep network features), and (3) an adversarial loss that balances a min-max game between a generator and a discriminator (standard GAN objective \cite{goodfellow2014generative}). The proposed framework basically favors outputs that are  perceptually similar to the high-dimensional images. To quantify this capability, they introduce a new Mean Opinion Score (MOS) which is assigned manually by human raters indicating bad/excellent quality of each super-resolved image. Since other techniques generally learn to optimize direct data dependent measures (such as pixel-errors), \cite{ledig2017photo} outperformed its competitors by a significant margin on the perceptual quality metric. 

\subsubsection{EnhanceNet} 
This network design focuses on creating faithful texture details in high-resolution super-resolved images \cite{sajjadi2017enhancenet}. A key problem with regular image quality measures such as PSNR is their noncompliance with the perceptual quality of an image. This results in overly smoothed images that do not have sharp textures. To overcome this problem, EnhanceNet used two other loss terms beside the regular pixel-level MSE loss: (a) the \emph{perceptual loss function} was defined on the intermediate feature representation of a pretrained network \cite{johnson2016perceptual} in the form of $\ell_1$ distance. (b) the \emph{texture matching loss} is used to match the texture of low and high resolution images and is quantified as the $\ell_1$ loss between gram matrices computed from deep features.  The whole network architecture is adversarialy trained where the SR network's goal is to fool a discriminator network. 

The architecture used by EnhanceNet is based on the Fully Convolutional Network~\cite{long2015fully} and residual learning principle \cite{kim2016VDSR}. Their results showed that although best PSNR is achieved when only a pixel level loss is used, the additional loss terms and an adversarial training mechanism lead to more realistic and perceptually better outputs. On the downside, the proposed adversarial training could create visible artifacts when super-resolving highly textured regions. This limitation was addressed further by the recent work on high perceptual quality SR~\cite{wang2018esrgan}.


\subsubsection{SRFeat}
~\cite{park2018srfeat} is another GAN-based Super-Resolution algorithm with Feature Discrimination.  This work focuses on the realistic perception of the input image using an additional discriminator that assists the generator to generate high-frequency structural features rather than noisy artifacts. This requisite is achieved by distinguishing between the features of synthetic (machine generated) and the real images. This network uses 9$\times$9 convolutional layer to extract features.  Then, residual blocks similar to \cite{he2016deep} with long-range skip connections are used which have 1$\times$1 convolutions. The feature maps are upsampled by pixel shuffler layers to achieve the desired output size. The authors used 16 residual blocks with two different settings of feature maps \ie 64 and 128. The proposed model uses a combination of perceptual (adversarial loss) and pixel-level loss ($\ell_2$) functions that is optimized with an Adam optimizer \cite{kingma2014adam}. The input resolution to the system is 74$\times$74 which only outputs 296$\times$296 image. The network uses 120k images from the ImageNet~\cite{deng2009imagenet} for pre-training the generator, followed by fine-tuning on augmented DIV2K dataset~\cite{timofte2017ntireFlicker2K} using learning rates of 10$^{-4}$ to 10$^{-6}$.

\subsubsection{ESRGAN}

Enhanced Super-Resolution Generative Adversarial Networks (ESRGAN) \cite{wang2018esrgan}  builds upon SRGAN \cite{ledig2017photo} by removing batch normalization and incorporating dense blocks. Each dense block's input is also connected to the output of the respective block making a residual connection over each dense block. ESRGAN also has a global residual connection to enforce residual learning.  Moreover, the authors also employ an enhanced discriminator called Relativistic GAN \cite{jolicoeur2018relativisticGAN}.

The training is performed on a total of 3,450 images from the DIV2K \cite{timofte2017ntireFlicker2K} and Flicker2K datasets employing augmentation \cite{timofte2017ntireFlicker2K} via the $\ell_1$ loss function first and then using the trained model using perceptual loss. The patch size for training is set to 128$\times$128, having a network depth of 23 blocks. Each block contains five convolutional layers, each with 64 feature maps. The visual results are comparatively better as compared to RCAN \cite{zhang2018RCAN}, however, it lags in terms of the quantitative measures where RCAN performs better.


\begin{figure}
    \centering
    \begin{minipage}{0.48\textwidth}
       \centering
       \includegraphics[width=\columnwidth]{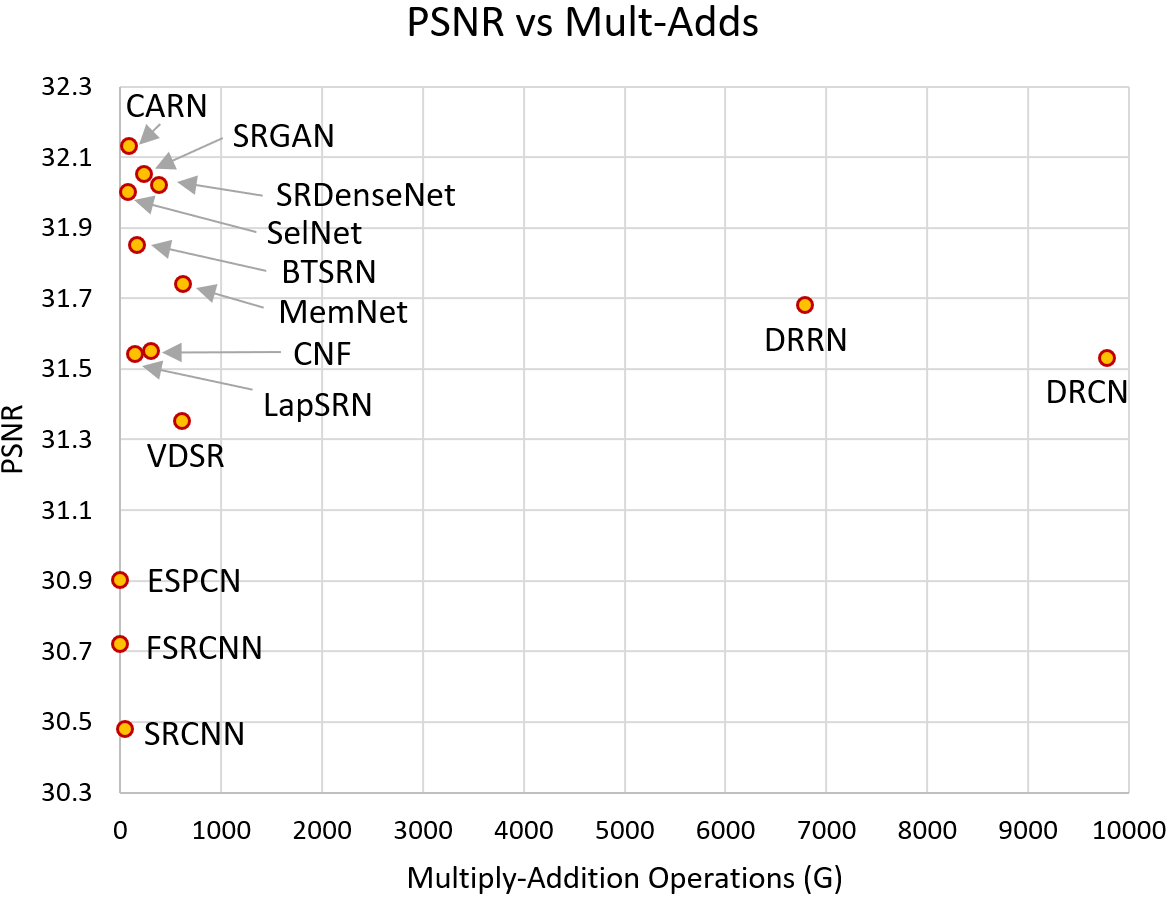}
       \caption{Comparison of Multiplication-Addition operations in various SR networks. Note that FLOPs are roughly double the number of mult-adds. Algorithmic runtime (during inference) is proportional to the multi-add operations.}
       \label{fig:flops}
    \end{minipage}\hfill
    \begin{minipage}{0.48\textwidth}
        \centering
        \includegraphics[width=\columnwidth]{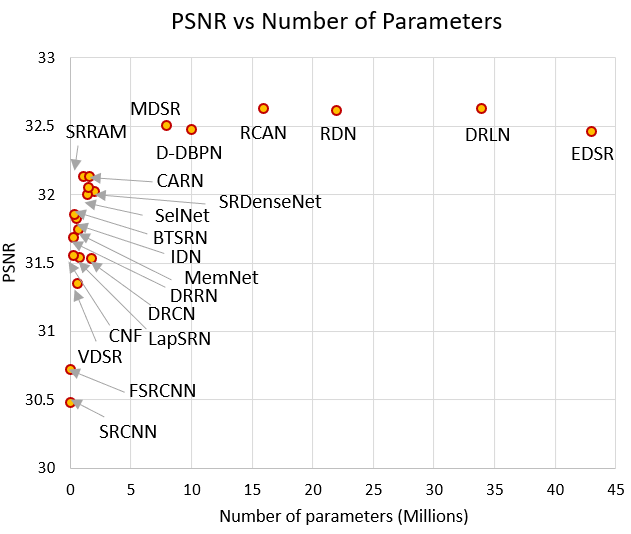}
        \caption{Comparison of number of parameters in various SR architectures. The memory footprint and training time of the model is directly related to the number of tunable parameters.}
       \label{fig:params}
    \end{minipage}
\end{figure}


\begin{table*}
\caption{Parameters comparison of CNN-based SR algorithms. GRL stands for Global residual learning, LRL means Local residual learning, MST is abbreviation of Multi-scale training.}
\centering
\begin{adjustbox}{max width=\textwidth}
\begin{tabular}{|l||ccccccccccc|}\hline 

Method      &Input      &Output  &Blocks    &Depth &Filters &Parameters &GRL   &LRL  &MST  & Framework      &Loss \\\hline \hline
SRCNN  		&bicubic    &Direct  &	        &3 	   &64 	    & 57k 	    &      &	 &	   & Caffe	        &$\ell_2$\\
FSRCNN 		&LR 		&Direct  &	        &8 	   &56 	    & 12k 	    &	   &	 &	   & Caffe          &$\ell_2$\\
ESPCN  		&LR 		&Direct  &	        &3 	   &64 	    & 20k 	    &	   &	 &	   & Theano         &$\ell_2$\\
SCN    		&bicubic    &Prog.   &\checkmark       &10	   &128	    & 42k  	    &	   &	 &	   & Cuda-CovNet    &$\ell_2$\\
REDNet      &bicubic    &Direct  &          &30    &128     & 4,131k    &\checkmark   &\checkmark  &     & Caffe          &$\ell_2$\\ 
VDSR   		&bicubic    &Direct  &	        &20	   &64 	    & 665k 	    &\checkmark   &\checkmark  &	   & Caffe	        &$\ell_2$\\
DRCN   		&bicubic    &Direct  & 	        &20	   &256	    & 1,775k	&\checkmark   &	 &	   & Caffe	        &$\ell_2$\\
LapSRN 		&LR 		&Prog.   &\checkmark       &24	   &64 	    & 812k 	    &\checkmark   &	 &     & MatConvNet     &$\ell_1$\\
DRRN   		&bicubic    &Direct  &\checkmark	    &52	   &128	    & 297k 	    &\checkmark   &\checkmark  &\checkmark  & Caffe          &$\ell_2$\\
SRGAN       &LR         &Direct  &\checkmark       &33    &64      & 1500k     &      &     &     & Theano/Lasagne &$\ell_2$\\
DnCNN       &bicubic    &Direct  &          &17    &64      & 566k      &      &     &\checkmark  & MatConvNet     &$\ell_2$\\
IRCNN       &bicubic    &Direct  &          &7     &64      & 188k      &      &     &\checkmark  & MatConvNet     &$\ell_2$\\
FormResNet  &bicubic    &Direct  &\checkmark       &20    &64      & 671k      &\checkmark   &     &\checkmark  & MatConvNet     &$\ell_2,\ell_{TV}$\\
EDSR        &LR         &Direct  &\checkmark       &65    &256     & 43000k    &\checkmark   &\checkmark  &     & Torch          &$\ell_1$\\
MDSR   		&LR 		&Direct  &\checkmark       &162   &64 	    & 8,000k	&\checkmark   &\checkmark  &\checkmark  & Torch          &$\ell_1$\\
ZSSR        &LR         &Direct  &          &8	   &64 	    & 225k 	    &\checkmark   &     &     & Tensorflow     &$\ell_1$\\
MemNet      &bicubic    &Direct  &\checkmark       &80    &64      & 677k      &\checkmark   &\checkmark  &\checkmark  & Caffe          &$\ell_2$\\ 
MS-LapSRN   &LR         &Prog.   &\checkmark       &84	   &64 	    & 222k 	    &\checkmark   &\checkmark  &\checkmark  & MatConvNet	    &$\ell_1$\\ 
CMSC        &bicubic    &Direct  &\checkmark       &35	   &64 	    & 1220k	    &\checkmark   &\checkmark  &\checkmark  & PyTorch        &$\ell_2$\\ 
CNF         &bicubic    &Direct  &          &15    &64	    & 337K 	    &      &     &     & Caffe          &$\ell_2$\\
IDN         &LR         &Direct  &\checkmark       &31    &64	    & 796k 	    &\checkmark   &\checkmark  &     & Caffe          &$\ell_2$,$\ell_1$\\
BTSRN       &LR         &Direct  &\checkmark       &22    &64	    & 410K 	    &\checkmark   &\checkmark  &     & Tensorflow	    &$\ell_2$\\
SelNet      &LR         &Direct  &          &22    &64	    & 974K 	    &\checkmark   &\checkmark  &     & MatConvNet	    &$\ell_2$\\
CARN        &LR         &Direct  &\checkmark       &32    &64	    & 1,592K 	&\checkmark   &\checkmark  &\checkmark  & PyTorch	    &$\ell_1$\\
SRMD        &LR         &Direct  &          &12	   &128 	& 1482k	    &      &     &     & MatConvNet	    &$\ell_2$\\ 
SRDenseNet  &LR         &Direct  &\checkmark       &64    &16-128  & 5,452k	&\checkmark   &\checkmark  &     & TensorFlow	    &$\ell_2$\\ 
EnhanceNet  &LR         &Direct  &\checkmark       &24	   &64 	    & 889k	    &      &\checkmark  &     & TensorFlow     &$\ell_2, \ell_t,GAN$\\ 
SRFeat      &LR         &Direct  &\checkmark       &54	   &128 	& 6,189k    &\checkmark   &\checkmark  &     & TensorFlow     &$\ell_2,\ell_p,GAN$\\
SRRAM       &LR         &Direct  &\checkmark       &64	   &64 	    & 1,090K    &\checkmark   &\checkmark  &\checkmark  & Tensorflow     &$\ell_1$\\
D-DBPN      &LR         &Direct  &\checkmark       &46	   &64 	    & 10,000K   &\checkmark   &\checkmark  &     & Caffe          &$\ell_2$\\ 
RDN         &LR         &Direct  &\checkmark       &149   &64 	    & 21,900k   &\checkmark   &\checkmark  &     & Torch          &$\ell_1$\\ 
ESRGAN      &LR         &Direct  &\checkmark       &115   &64	    & 38,549k	&\checkmark   &\checkmark  &     & Pytorch	    &$\ell_1$\\
SRFBN & LR & Direct & \checkmark  & 28 & 64 & 3,500k &  \checkmark & \checkmark & \checkmark & Pytorch & $\ell_1$\\
RCAN        &LR         &Direct  &\checkmark       &500   &64      & 16,000k  	&\checkmark   &\checkmark  &\checkmark  & Pytorch	    &$\ell_1$\\
DRLN        &LR         &Direct  &\checkmark       &160   &64      & 34,000k  	&\checkmark   &\checkmark  &\checkmark  & Pytorch	    & $\ell_1$\\
EBRN & LR & Direct & \checkmark & 173 & 64 & 7,900k & & \checkmark &  & Pytorch & $\ell_1$, $\ell_2$ \\ \hline
\end{tabular}
\end{adjustbox}
\label{table:parameters}
\end{table*}

\section{Experimental Evaluation}

\subsection{Datasets}
In this section, we compare the state-of-the-art algorithms on publicly available benchmark datasets which include Set5~\cite{bevilacqua2012Set5}, Set14~\cite{zeyde2010Set14}, BSD100~\cite{martin2001BSD100}, Urban100~\cite{huang2015URBAN100}, DIV2K~\cite{timofte2017ntireFlicker2K} and Manga109~\cite{fujimoto2016manga109}. The representative images from all the datasets are shown in Figure~\ref{fig:dataset_images}.
\begin{itemize}
\item \textbf{Set5} \cite{bevilacqua2012Set5} is a classical dataset and only contains five test images of a baby, bird, butterfly, head, and a woman.

\item  \textbf{Set14} \cite{zeyde2010Set14} consists of more categories as compared to Set5~\cite{bevilacqua2012Set5}; however, the number of images are still low \ie 14 test images. 

\item  \textbf{BSD100} \cite{martin2001BSD100} is another classical dataset having 100 test images proposed by Martin \etal \cite{martin2001BSD100}. The dataset is composed of a large variety of images ranging from natural images to object-specific such as plants, people, food \etc

\item  \textbf{Urban100} \cite{huang2015URBAN100} is a relatively more recent dataset introduced by Huang \etal The number of images is the same as BSD100~\cite{martin2001BSD100}; however, the composition is entirely different. The focus of the photographs is on human-made structures \ie urban scenes.

\item  \textbf{DIV2K} \cite{timofte2017ntireFlicker2K} is a dataset used for NITRE challenge. The image quality is of 2K resolution and is composed of 800 images for training while 100 images each for testing and validation. As the test set is not publicly available, the results are only reported on validation images for all the algorithms. 

\item  \textbf{Manga109} \cite{fujimoto2016manga109} is the latest addition for evaluating super-resolution algorithms. The dataset is a collection of 109 test images of a manga volume. These mangas were professionally drawn by Japanese artists and were available only for commercial use between the 1970s and 2010s. 

\end{itemize}


\subsection{Quantitative Measures}

The algorithms detailed in section~\ref{sec:sisr} were evaluated on the peak signal-to-noise ratio (PSNR) and the structural similarity index (SSIM)~\cite{Wang2004} measures. Table~\ref{table:Results_benchmark} presents the results for 2$\times$ and 3$\times$ while Table~\ref{table:Results_benchmark_4x} is for 4$\times$ the super-resolution algorithms. 
Currently, the PSNR and SSIM performance of DRLN~\cite{anwar2019DRLN} is better for 2$\times$ and 3$\times$ and ESRGAN~\cite{wang2018esrgan} for 4$\times$. However, it is difficult to declare one algorithm to be a clear winner compared to the rest as there are many factors involved such as network complexity, depth of the network, training data, patch size for training, number of features maps, \etc A fair comparison is only possible by keeping all the parameters consistent.

\begin{figure*}
\begin{center}
\begin{tabular}{c@{ } c@{ }  c@{ } c@{ } c@{ } c}
    
    \multirow{4}{*}{\includegraphics[width=.24\textwidth,height=4.45cm,valign=t]{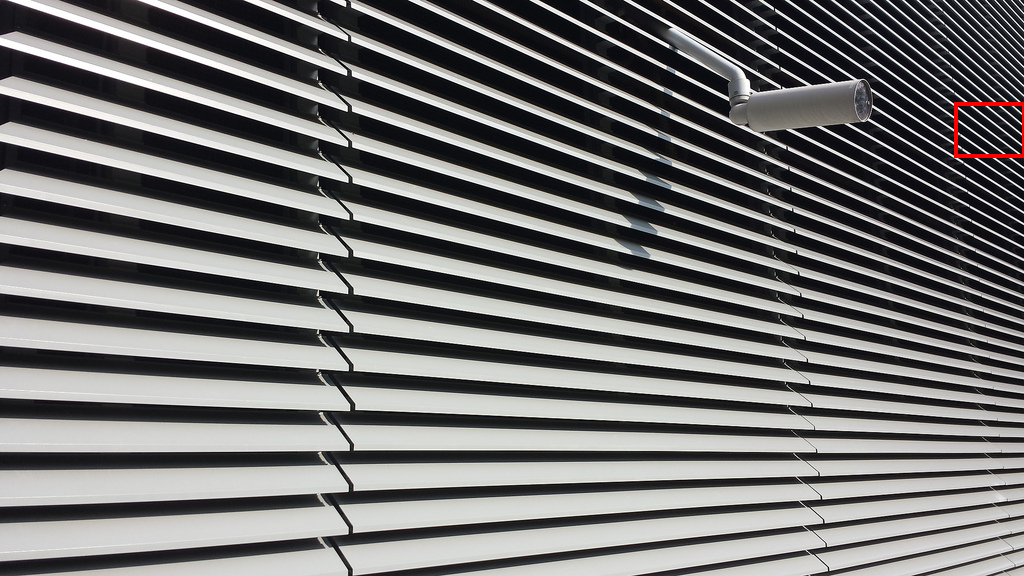}} &  
    \includegraphics[width=.145\textwidth,valign=t]{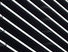}&
  	\includegraphics[width=.145\textwidth,valign=t]{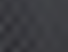}&   
    \includegraphics[width=.145\textwidth,valign=t]{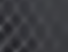}&   
 	\includegraphics[width=.145\textwidth,valign=t]{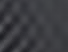}&
  	\includegraphics[width=.145\textwidth,valign=t]{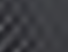}\\
    & Original  & Bicubic       & SRCNN~\cite{dong2016SRCNNPAMI}         &  FSRCNN~\cite{dong2016FSRCNN}         & VDSR~\cite{kim2016VDSR}\\

    &
    \includegraphics[width=.145\textwidth,valign=t]{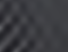}&
    \includegraphics[width=.145\textwidth,valign=t]{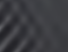}&  
    \includegraphics[width=.145\textwidth,valign=t]{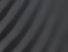}&    
    \includegraphics[width=.145\textwidth,valign=t]{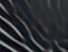}&
    \includegraphics[width=.145\textwidth,valign=t]{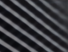}\\
    URBAN~\cite{huang2015URBAN100} (8$\times$) & DRCN~\cite{kim2016DRCN}         & DRRN~\cite{tai2017DRRN}&MSLapSRN~\cite{MSLapSRN}      & RCAN~\cite{zhang2018RCAN}          & DRLN~\cite{anwar2019DRLN}\\

    \multirow{4}{*}{\includegraphics[width=.24\textwidth,trim={0cm 10.3cm 0cm 1cm},clip,height=3.2cm,valign=t]{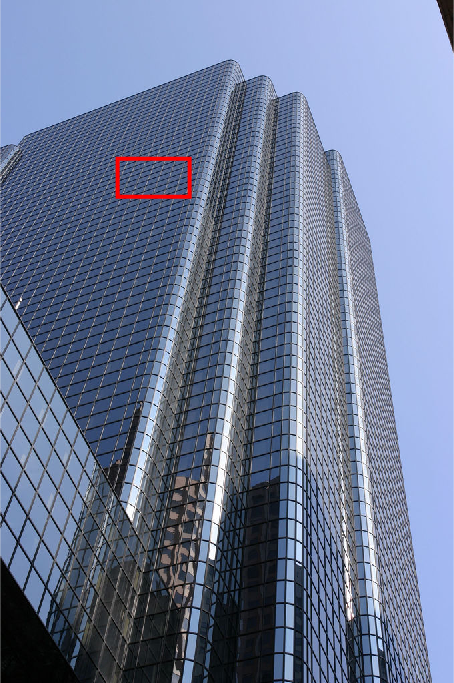}} &  
    \includegraphics[width=.145\textwidth,valign=t]{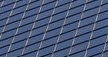}&
  	\includegraphics[width=.145\textwidth,valign=t]{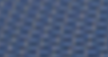}&   
    \includegraphics[width=.145\textwidth,valign=t]{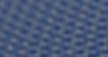}&   
 	\includegraphics[width=.145\textwidth,valign=t]{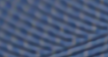}&
  	\includegraphics[width=.145\textwidth,valign=t]{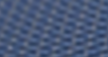}\\
    & Original  & Bicubic       & SRCNN~\cite{dong2016SRCNNPAMI}         &  IRCNN~\cite{zhang2017IrCNN}         & VDSR~\cite{kim2016VDSR}\\

    &
    \includegraphics[width=.145\textwidth,valign=t]{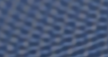}&
    \includegraphics[width=.145\textwidth,valign=t]{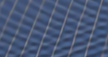}&  
    \includegraphics[width=.145\textwidth,valign=t]{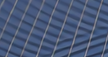}&    
    \includegraphics[width=.145\textwidth,valign=t]{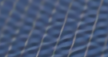}&
    \includegraphics[width=.145\textwidth,valign=t]{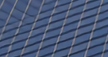}\\
    URBAN~\cite{huang2015URBAN100} (4$\times$) & MSLapSRN~\cite{MSLapSRN}         & EDSR~\cite{lim2017EDSR}& RCAN~\cite{zhang2018RCAN}          & CARN~\cite{ahn2018CARN}      &DRLN~\cite{anwar2019DRLN}\\
    
\end{tabular}
\end{center}
\caption{Super-resolution comparison on 8$\times$ and 4$\times$ sample images with sharp edges and texture, taken from URBAN100~\cite{huang2015URBAN100}.}
\label{fig:CNN_images}
\end{figure*}

\begin{figure*}
\begin{center}
\begin{tabular}{c@{ }c@{ }c@{ }c@{ }c@{ }c@{ }c}
    &
     \multirow{2}{*}{\includegraphics[width=.24\textwidth,trim={0.2cm 0cm 0cm 0cm},clip,valign=t]{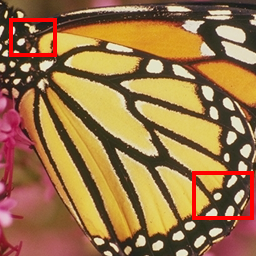}} &  
  	\includegraphics[width=.145\textwidth,valign=t]{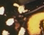}&   
    \includegraphics[width=.145\textwidth,valign=t]{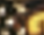}&  
 	\includegraphics[width=.145\textwidth,valign=t]{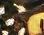}&
 	\includegraphics[width=.145\textwidth,valign=t]{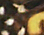}&
  	\includegraphics[width=.145\textwidth,valign=t]{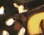}\\
   
    \rotatebox{90}{Set5}& &
  	\includegraphics[width=.145\textwidth,valign=t]{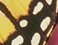}&   
    \includegraphics[width=.145\textwidth,valign=t]{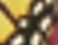}&  
 	\includegraphics[width=.145\textwidth,valign=t]{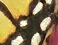}&
 	\includegraphics[width=.145\textwidth,valign=t]{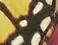}&
  	\includegraphics[width=.145\textwidth,valign=t]{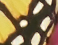}\\
  	\\
  	  &
     \multirow{2}{*}{\includegraphics[width=.24\textwidth,trim={0.0cm 2.8cm 0cm 0cm},clip,valign=t]{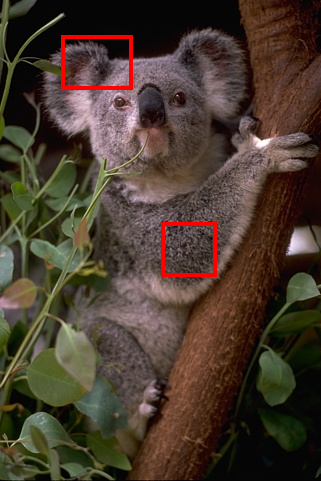}} &  
  	\includegraphics[width=.145\textwidth,valign=t]{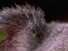}&   
    \includegraphics[width=.145\textwidth,valign=t]{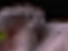}&  
 	\includegraphics[width=.145\textwidth,valign=t]{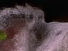}&
 	\includegraphics[width=.145\textwidth,valign=t]{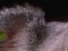}&
  	\includegraphics[width=.145\textwidth,valign=t]{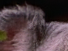}\\
   
    \rotatebox{90}{BSD100}& &
  	\includegraphics[width=.145\textwidth,valign=t]{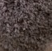}&   
    \includegraphics[width=.145\textwidth,valign=t]{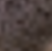}&  
 	\includegraphics[width=.145\textwidth,valign=t]{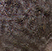}&
 	\includegraphics[width=.145\textwidth,valign=t]{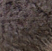}&
  	\includegraphics[width=.145\textwidth,valign=t]{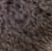}\\
  	\\
  	
  	    &
     \multirow{2}{*}{\includegraphics[width=.24\textwidth,trim={0.0cm 4cm 0cm 4cm},clip,valign=t]{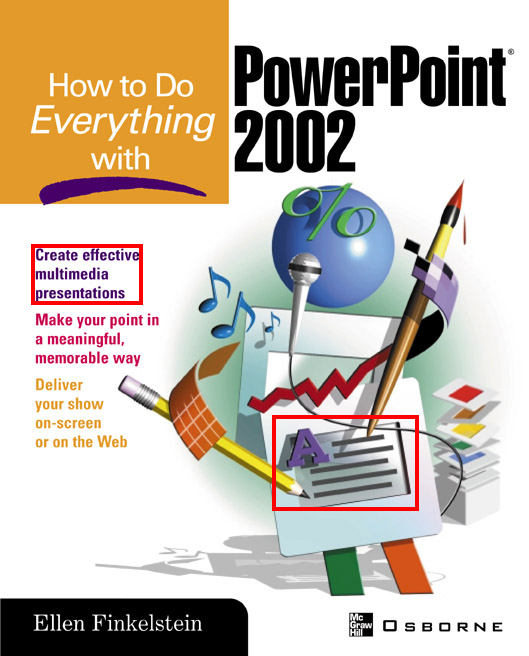}} &  
  	\includegraphics[width=.145\textwidth,valign=t]{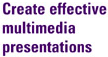}&   
    \includegraphics[width=.145\textwidth,valign=t]{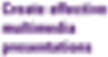}&  
 	\includegraphics[width=.145\textwidth,valign=t]{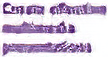}&
 	\includegraphics[width=.145\textwidth,valign=t]{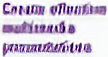}&
  	\includegraphics[width=.145\textwidth,valign=t]{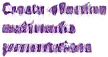}\\
   
    \rotatebox{90}{Set14}& &
  	\includegraphics[width=.145\textwidth,valign=t]{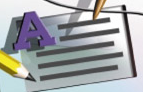}&   
    \includegraphics[width=.145\textwidth,valign=t]{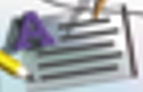}&  
 	\includegraphics[width=.145\textwidth,valign=t]{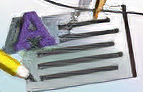}&
 	\includegraphics[width=.145\textwidth,valign=t]{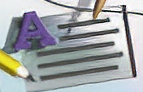}&
  	\includegraphics[width=.145\textwidth,valign=t]{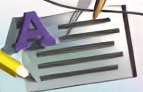}\\
    && Ground-truth & Bicubic & EnhanceNet& SRGAN &ESRGAN \\
 \end{tabular}
\end{center}
\caption{Qualitative comparison for generative adversarial network algorithms for 4$\times$ super-resolution.}
\label{fig:GAN_images}
\end{figure*}

In Figure~\ref{fig:CNN_images}, we present the visual comparison between a few of the state-of-the-art algorithms which aim to improve the PSNR of the images. Furthermore, Figure~\ref{fig:GAN_images} shows the output of the GAN-based algorithms which are perceptually-driven and aim to enhance the visual quality of the generated outputs. As one can notice, outputs in Figure~\ref{fig:GAN_images} are generally more crisp, but the corresponding PSNR values are relatively lower compared to methods that optimize pixel-level loss measures. 

\begin{table*}
\caption{Mean PSNR and SSIM for the SR methods evaluated on the benchmark datasets. The '-' indicates that the method is not suitable to handle the images of the corresponding dataset or used the dataset during training or the source code is not available publicly.}
\centering
\resizebox{\textwidth}{!}{
\begin{tabular}{|l||l||cc||cc||cc||cc||cc||cc|}\hline
  &    &  \multicolumn{2}{c}{Set5}	  	    &  \multicolumn{2}{c}{Set14}	   &  \multicolumn{2}{c}{BSD100}	 &	\multicolumn{2}{c}{Urban100}	 & \multicolumn{2}{c|}{DIV2K}	& \multicolumn{2}{c|}{Manga109} \\ \hline \hline
Scale & Method & PSNR & SSIM & PSNR & SSIM  & PSNR & SSIM & PSNR & SSIM & PSNR & SSIM & PSNR & SSIM\\ \hline
& Bicubic& 33.68 &0.9304 &30.24 	&0.8691 &29.56 	&0.8435	&26.88	&0.8405	&32.45 	& 0.904 &31.05 & 0.935\\
&SRCNN          & 36.66 &0.9542 &32.45	&0.9067	&31.36	&0.8879 &29.51	&0.8946	&34.59 	& 0.932 &35.72& 0.968\\
&FSRCNN         & 36.98	&0.9556	&32.62	&0.9087	& 31.50	&0.8904	&29.85	&0.9009	&34.74 	& 0.934 &36.62 &0.971\\
&SCN            & 36.52 &0.953 	&32.42 	&0.904 	&31.24 	&0.884 	&29.50 	&0.896 	&34.98	& 0.937 &35.51 & 0.967\\
&REDNet         & 37.66	&0.9599	&32.94	&0.9144	&31.99	&0.8974	&-		&	-	&	-	&-		&-      &-\\
&VDSR 			& 37.53 &0.9587	&33.05	&0.9127	& 31.90	&0.8960 &30.77	&0.9141 &35.43 	& 0.941 &37.16 &0.974\\
&DRCN    		& 37.63 &0.9588 &33.06 	&0.9121 &31.85 	&0.8942 &30.76  &0.9133 &35.45 	& 0.940	&37.57 &0.973\\
&LapSRN  		& 37.52 &0.9591 &32.99 	&0.9124 &31.80 	&0.8949 &30.41  &0.9101 &35.31 	& 0.940 &37.53 &0.974\\
&DRRN    							& 37.74 &0.9591 &33.23 	&0.9136 &32.05 	&0.8973 &31.23  &0.9188 &35.63 	& 0.941 &37.92 &0.976\\
&DnCNN								& 37.58	&0.9590	&33.03	&0.9128 &31.90	&0.8961	& 30.74	&0.9139 &	-	&-		&-      &-\\
&EDSR                               &38.11 &0.9602 &33.92 &0.9195 &32.32 &0.9013 &32.93 &0.9351& 35.03 &0.9695  &39.10   &0.9773\\
&MDSR                               &38.11 &0.9602  &33.85 &0.9198 &32.29 &0.9007 &32.84 &0.9347 &34.96 &0.9692 &38.96   &0.978\\
&ZSSR 								& 37.37 &0.9570 &33.00 	&0.9108 &31.65 &0.8920	&  -		&  -		& - 		&  -		&-&-\\
&MemNet  							& 37.78 &0.9597 &33.28 	&0.9142 &32.08 	&0.8978 &31.31 	&0.9195 &	-	&	-	&37.72 &0.9740\\
&CMSC								& 37.89 &0.9605 &33.41 	&0.9153 &32.15 	&0.8992 &31.47 	&0.9220 &-		&	-	&&\\
&IDN                                &37.83  &0.9600 &33.30  &0.9148 &32.08  &0.8985 &31.27  &0.9196        &-&-&38.02&0.9749\\
&CNF                                &37.66  &0.9590 &33.38 &0.9136 &31.91 &0.8962   &-     &  -      &-&-&-&-\\
&BTSRN                              &37.75  &-      &33.20 &-      &32.05 &-        &31.63 &-       &-&-&-&-\\
&SRMDNF 							& 37.79 &0.9601	&33.32 	&0.9159 &32.05	&0.8985	&31.33 	&0.9204	&35.54		& 0.9414		&38.07 &0.9761\\
&D-DBPN  							& 38.09 &0.9600 &33.85 	&0.9190 &32.27 	&0.9000 &32.55 	&0.9324	&	-	&	-	&38.89 &0.9775\\
&SelNet                             &37.89  &0.9598 &33.61 &0.9160 &32.08 &0.8984   &-     &    -    &-&-&-&-\\
&CARN                               &37.76  &0.9590 &33.52 &0.9166 &32.09 &0.8978   &31.92 &0.9256  &36.04  &0.9451  &38.36 &0.9764\\
&SRRAM                                &37.82 & 0.9592 &33.48 & 0.9171 &32.12 & 0.8983 &32.05 & 0.9264 &  -     &   -     &    -  &\\
&RDN 								& 38.24 &0.9614 &34.01	&0.9212 &32.34 	&0.9017 &32.89	&0.9353	&	-	&	-	 &39.18 & 0.9780\\
&  SRFBN & 38.11 & 0.9609 & 33.82 & 0.9196 & 32.29 & 0.9010 & 32.62 & 0.9328  & - & - & 39.08 & 0.9779\\
\multirow{-24}{*}{$\times$2}&RCAN                               &38.27 &0.9614  &34.12  &0.9216 &32.41  &0.9027 &33.34 &0.9384&	36.63	& 0.9491 &39.44 &0.9786\\
&DRLN                               &38.27 &0.9616 &34.28 &0.9231 &32.44 &0.9028 &33.37 &0.9390 &	-	&	- &39.58 &0.9786\\
& EBRN & 38.35 & 0.9620 & 34.24 & 0.9226 & 32.47 & 0.9033 & 33.52 & 0.9402 & - & - & 39.62 & 0.9802\\

\hline \hline

&Bicubic & 30.40 &0.8686 &27.54  &0.7741 &27.21 &0.7389 	&24.46 &0.7349 &29.66 	& 0.831 &26.95 &0.856\\
&SRCNN   							& 32.75 &0.9090 &29.29 	&0.8215 &28.41 &0.7863 	&26.24 &0.7991 &31.11 	& 0.864 &30.48 &0.912\\
&FSRCNN  							& 33.16 &0.9140 &29.42 	&0.8242 &28.52 &0.7893 	&26.41 &0.8064 &31.25 	& 0.868 &31.10 &0.921\\
&SCN 								& 32.62 & 0.908 &29.16 	& 0.818 &28.33 & 0.783 	&26.21 & 0.801 &31.42 	& 0.870 &30.22 & 0.914\\
&REDNet								&33.82	&0.9230	&29.61	&0.8341	&28.93 &0.7994	&-	   &	-   &-		&	-	&-      &-\\
&VDSR    							& 33.66 &0.9213 &29.78  &0.8318 &28.83 &0.7976  &27.14 &0.8279 & 31.76 & 0.878  &32.01 &0.934\\
&DRCN    							& 33.82 &0.9226 &29.77  &0.8314 &28.80 &0.7963  &27.15 &0.8277 & 31.79 & 0.877  &32.31 &0.936\\
&LapSRN  							& 33.82 &0.9227 &29.79  &0.8320 &28.82 &0.7973  &27.07 &0.8271 & 31.22 & 0.861  &32.21 &0.935\\
&DRRN    							& 34.03 &0.9244 &29.96  &0.8349 &28.95 &0.8004  &27.53 &0.8377 &31.96  & 0.880  &32.74 &0.939\\
&DnCNN								&33.75  &0.9222 & 29.81 &0.8321 &28.85 &0.7981	&27.15 &0.8276 &	-	&	-	&-&-\\
&EDSR                               &34.65 &0.9280 &30.52 &0.8462 &29.25 &0.8093 &28.80 &0.8653 &31.26 & 0.9340 &34.17 &0.9476\\
&MDSR                               &34.66 &0.9280  &30.44 &0.8452 &29.25 &0.8091 &28.79 &0.8655  &31.25    &0.9338 &34.17  &0.947\\
&ZSSR 								&33.42  &0.9188 & 29.80 &0.8304 &28.67 & 0.7945 &  	 -  & - 	   & - 		& - 		&   -    &-\\
&MemNet  							&34.09  &0.9248 &30.00  &0.8350 &28.96 &0.8001  &27.56 &0.8376 &-		&-		&32.51  &0.9369\\
&CMSC    							&34.24  &0.9266 &30.09  &0.8371 &29.01 &0.8024  &27.69 &0.8411 &-		&-		&  -     &-\\ 
&IDN                                &34.11  &0.9253 &29.99  &0.8354 &28.95 &0.8013  &27.42 &0.8359 &-        & -      &32.69  &0.9378\\
&CNF                                &33.74&0.9226 &29.90&0.8322 &28.82&0.7980 &-&    -       &-&-&-&- \\
&BTSRN                              &34.03&-      &29.90&-      &28.97&-      &27.75&-      &-&-&-&-\\
&SRMDNF 							&34.12  &0.9254 &30.04  &0.8382 &28.97 &0.8025	&27.57 &0.8398 & 31.92	&0.8801	&33.00 &0.9403\\
&SelNet                             &34.27&0.9257 &30.30&0.8399 &28.97&0.8025 &-        & -     &-&-&-&-\\
&CARN                               &34.29&0.9255 &30.29&0.8407 &29.06&0.8034 &28.06&0.8493 &32.37 &0.8871 &33.49& 0.9440\\
&SRRAM                                &34.30 &0.9256 &30.32 &0.8417 &29.07 &0.8039 &28.12&0.8507&-&-&-&-\\
&RDN 								&34.71  &0.9296 &30.57  &0.8468 &29.26 &0.8093	&28.80 &0.8653 &-        & -		&34.13 &0.9484\\
 & SRFBN & 34.70 & 0.9292 & 30.51 & 0.8461 & 29.24 & 0.8084 & 28.73 & 0.8641 & 34.18 & 0.9481 \\
\multirow{-23}{*}{$\times$3}&RCAN                               &34.74  &0.9299 &30.65  &0.8482 &29.32 &0.8111 &29.09 &0.8702 &	32.80	& 0.8941  &34.44 &0.9499\\
&DRLN &34.78 &0.9303 &30.73 &0.8488 &29.36 &0.8117 &29.21 &0.8722  &-        & - &34.71 &0.9509\\
\hline 
\end{tabular}
}
\label{table:Results_benchmark}
\end{table*}

\begin{table*}
\caption{Mean PSNR and SSIM for the SR methods evaluated on the benchmark datasets for a higher super-resolution factor 4$\times$.}
\centering
\resizebox{\textwidth}{!}{
\begin{tabular}{|l||l||cc||cc||cc||cc||cc||cc|}\hline
  &    &  \multicolumn{2}{c}{Set5}	  	    &  \multicolumn{2}{c}{Set14}	   &  \multicolumn{2}{c}{BSD100}	 &	\multicolumn{2}{c}{Urban100}	 & \multicolumn{2}{c|}{DIV2K}	& \multicolumn{2}{c|}{Manga109} \\ \hline \hline
Scale & Method & PSNR & SSIM & PSNR & SSIM  & PSNR & SSIM & PSNR & SSIM & PSNR & SSIM & PSNR & SSIM\\ \hline

&Bicubic &28.43 &0.8109 &26.00 	&0.7023 &25.96 	&0.6678 &23.14 &0.6574  &28.11 	& 0.775	&25.15 & 0.789\\
&SRCNN   							&30.48 &0.8628 &27.50 	&0.7513 &26.90 	&0.7103 &24.52 &0.7226  &29.33 	& 0.809 &27.66 & 0.858\\
&FSRCNN  							&30.70 &0.8657 &27.59 	&0.7535 &26.96 	&0.7128 &24.60 &0.7258  &29.36 	& 0.811 &27.89 & 0.859\\
&SCN 								&30.39 &0.862  &27.48 	&0.751  &26.87 	& 0.710 &24.52 &0.725   &29.47	& 0.813 &27.39 & 0.857\\
&REDNet								&31.51 &0.8869 &27.86	&0.7718 &27.40	&0.7290	&-	   &-	    &-		&-		&-     &-\\
&VDSR    							&31.35 &0.8838 &28.02 	&0.7678 &27.29 	&0.7252 &25.18 &0.7525  &29.82 	& 0.824 &28.82 &0.886\\
&DRCN    							&31.53 &0.8854 &28.03 	&0.7673 &27.24 	&0.7233 &25.14 &0.7511  &29.83 	& 0.823 &28.97 &0.886\\
&LapSRN  							&31.54 &0.8866 &28.09 	&0.7694 &27.32 	&0.7264 &25.21 &0.7553  &29.88 	& 0.825 &29.09 &0.890\\
&DRRN    							&31.68 &0.8888 &28.21 	&0.7720 &27.38 	&0.7284 &25.44 &0.7638  &29.98 	& 0.827 &29.46 &0.896\\
&SRGAN								&32.05 &0.8910 &28.53 	&0.7804 &27.57 	&0.7354	&26.07 &0.7839  &28.92	& 0.896	&-     &-\\
&DnCNN								&31.40 &0.8845 &28.04 	&0.7672 &27.29	&0.7253	&25.20 &0.7521   &-		&-		&-     &-\\
&EDSR                               &32.46 &0.8968 &28.80   &0.7876 &27.71  &0.7420 &26.64 &0.8033  &29.25  & 0.9017 &31.02 &0.9148\\
&MDSR                               &32.50 &0.8973 &28.72   &0.7857 &27.72  &0.7418 &26.67 &0.8041 &29.26   &0.9016  &31.11 &0.915\\
&ZSSR 								&31.13 &0.8796 &28.01 	&0.7651 &27.12 	&0.7211 &-	   &-	    &-		&-		&-     &-\\
&MemNet  							&31.74 &0.8893 &28.26 	&0.7723 &27.40 	&0.7281 &25.50 	&0.7630 &-		&-		&29.42  &0.8942\\
&CMSC    							&31.91 &0.8923 &28.35 	&0.7751 &27.46 	&0.7308 &25.64 	&0.7692  &-		&-		&-     &-\\
&IDN                                & 31.82 &0.8903 &28.25 &0.7730 &27.41 &0.7297 &25.41 &0.7632 && & 29.40 & 0.8936\\
&BTSRN                              &31.82 &0.8903 &28.25   &0.7730 &27.41  &0.7297 &25.41  &0.7632  &-		&-		&-     &-\\
&SRMDNF 							&31.96 &0.8925 &28.35 	&0.7787 &27.49 	&0.7337 &25.68 	&0.7731 &30.01	&0.8278	&30.09 &0.9024\\
&D-DBPN 							&32.47 &0.8980 &28.82 	&0.7860 &27.72	&0.7400 &26.38 	&0.7946  &-		&-		&30.91 &0.9137\\
&CNF                                &31.55 &0.8856 &28.15   &0.7680 &27.32  &0.7253 &-	   &-	    &-		&-		&-     &-\\
&BTSRN                              &31.85 &-      &28.20   &-      &27.47  &-      &25.74  &-      &-		&-		&-     &-\\
&SelNet                             &32.00 &0.8931 &28.49   &0.7783 &27.44  &0.7325 &-	   &-	    &-		&-		&-     &-\\
&CARN                               &32.13 &0.8937 &28.60   &0.7806 &27.58  &0.7349 &26.07  &0.7837 & 30.43 &0.8374 &30.40  & 0.9082\\
&SRRAM                                &32.13 &0.8932 &28.54   &0.7800 &27.56  &0.7350 &26.05  &0.7834 &-		&-		&-     &-\\
&SRDenseNet 	                    &32.02 &0.8934 &28.50	&0.7782 &27.53  &0.7337	& 26.05 &0.7819 &-		&-		&-     &-\\
&RDN    							&32.47 &0.8990 &28.81 	&0.7871	&27.72	&0.7419 &26.61 	&0.8028  &-		&-		&31.00 &0.9151\\
&ESRGAN                             &32.73 &0.9011 &28.99   &0.7917 &27.85  &0.7455 &27.03  &0.8153 &-		&-	 &31.66&0.9196\\
& SRFBN & 32.47 & 0.8983  & 28.81 & 0.7868 & 27.72 & 0.7409 & 26.60 & 0.8015 & - & - & 31.15 & 0.9160\\
\multirow{-28}{*}{$\times$4}&RCAN                               &32.63 &0.9002 &28.87   &0.7889 &27.77  &0.7436 &26.82  &0.8087 &30.77	& 0.8459&31.22 &0.9173\\
&DRLN & 32.63 &0.9002 &28.94 &0.7900 &27.83 &0.7444 &26.98 &0.8119 &-		&- &31.54 &0.9196\\
& EBRN & 32.79 &  0.9032 &  29.01 &  0.7903 & 27.85 &  0.7464 & 27.03 &  0.8114 & - & - &  31.53 &  0.9198 \\
\hline
\end{tabular}
}
\label{table:Results_benchmark_4x}
\end{table*}

\subsection{8$\times$ Super-resolution}
Most of the algorithms are generally evaluated on the standard datasets up to 4$\times$ super-resolution. When we tested these algorithms for higher magnification levels, the artifacts in the images became more visible (in table~\ref{table:Results_8x} and Figure~\ref{fig:CNN_images} the comparisons are provided for 8$\times$ super-resolution).  It is clear from the images that most of the state-of-the-art algorithms struggle to reproduce the textures in high magnified versions of the images.

\begin{table*}[!t]
\caption{The performance of state-of-the-art algorithms on widely used publicly available datasets, in terms of PSNR (in dB) and SSIM for 8$\times$.}
\centering
\resizebox{\textwidth}{!}{
\begin{tabular}{|l|l|cc||cc||cc||cc||cc|}
\hline
Scale & Method &\multicolumn{2}{c||} {SET5~\cite{bevilacqua2012Set5}}
& \multicolumn{2}{c||} {SET14~\cite{zeyde2010Set14}}
& \multicolumn{2}{c||} {BSD100~\cite{martin2001BSD100}}
& \multicolumn{2}{c||} {URBAN100~\cite{huang2015URBAN100}}
& \multicolumn{2}{c} {MANGA109~\cite{fujimoto2016manga109}}\\ \hline \hline
         &   &PSNR  &SSIM   &PSNR  &SSIM   &PSNR  &SSIM   &PSNR  &SSIM   &PSNR  &SSIM\\ \hline \hline
         
&Bicubic   &24.40 &0.6580 &23.10 &0.5660 &23.67 &0.5480 &20.74 &0.5160 &21.47 &0.6500\\
&SRCNN     &25.33 &0.6900 &23.76 &0.5910 &24.13 &0.5660 &21.29 &0.5440 &22.46 &0.6950\\
&FSRCNN    &20.13 &0.5520 &19.75 &0.4820 &24.21 &0.5680 &21.32 &0.5380 &22.39 &0.6730\\
&SCN       &25.59 &0.7071 &24.02 &0.6028 &24.30 &0.5698 &21.52 &0.5571 &22.68 &0.6963\\
&VDSR      &25.93 &0.7240 &24.26 &0.6140 &24.49 &0.5830 &21.70 &0.5710 &23.16 &0.7250\\
&LapSRN    &26.15 &0.7380 &24.35 &0.6200 &24.54 &0.5860 &21.81 &0.5810 &23.39 &0.7350\\
8$\times$  &MemNet  &26.16 &0.7414 &24.38 &0.6199 &24.58 &0.5842 &21.89 &0.5825 &23.56 &0.7387\\
&MSLapSRN  &26.34 &0.7558 &24.57 &0.6273 &24.65 &0.5895 &22.06 &0.5963 &23.90 &0.7564\\
&EDSR      &26.96 &0.7762 &24.91 &0.6420 &24.81 &0.5985 &22.51 &0.6221 &24.69 &0.7841\\
&D-DBPN    &27.21 &0.7840 &25.13 &0.6480 &24.88 &0.6010 &22.73 &0.6312 &25.14 &0.7987\\
&RCAN      &27.31 &0.7878 &25.23 &0.6511 &24.98 &0.6058 &23.00 &0.6452 &25.24 &0.8029\\
&DRLN      &27.36 & 0.7882 & 25.34 &0.6531 & 25.01 &0.6057 & 23.06 & 0.6471 & 25.29 &0.8041\\
& EBRN & 27.45 &  0.7908  & 25.44 & 0.6542 & 25.12 &  0.6079 & 23.32 & 0.6498 & 25.51 & 0.8085\\
\hline
\end{tabular}}
\label{table:Results_8x}
\end{table*}

\subsection{Number of parameters}
Table \ref{table:parameters} shows the comparison of parameters for different SR algorithms. Methods with direct reconstruction perform one-step upsampling from the LR to HR space, while progressive reconstruction predicts HR images in multiple upsampling steps. Depth represents the number of convolutional and transposed convolutional layers in the longest path from input to output for 4$\times$ SR. Global residual learning (GRL) indicates that the network learns the difference between the ground truth HR image and the upsampled (\ie using bicubic interpolation or learned filters) LR images. Local residual learning (LRL) stands for the local skip connections between intermediate convolutional layers. As one can notice, methods that perform  late upsampling \cite{dong2016FSRCNN,shi2016ESPCN} have considerably lower computational cost compared to methods that perform upsampling earlier in the network pipeline \cite{lim2017EDSR,dong2014SRCNN,zhang2018RCAN}.

\subsection{Choice of network loss}
\label{ss:networkloss}
The most popular choices for network loss is either mean square error $\ell_2$ or mean absolute error $\ell_1$ in the convolutional neural network for the image super-resolution. Similarly, Generative adversarial networks (GANs) also employ perceptual loss (adversarial loss) in 
addition to the pixel-level losses such as the MSE. From 
Table~\ref{table:parameters}, it is evident that the initial CNN methods were trained using $\ell_2$ loss; however, there is a shift in the trend towards $\ell_1$ more recently, and absolute mean difference measure  ($\ell_1$) has shown to be more robust compared to $\ell_2$. The reason is that $\ell_2$ puts more emphasis on more erroneous predictions while $\ell_1$ considers a more balanced error distribution.

\subsection{Network depth}
Contrary to the claim made in SRCNN~\cite{dong2016SRCNNPAMI} that network depth does not contribute to the better numbers rather it sometimes degrades the quality, VDSR~\cite{kim2016VDSR} initially proved that using deeper networks helps in better PSNR and image quality. EDSR~\cite{lim2017EDSR} further established this claim, where the number of convolutional layers were increased by nearly four times that of VDSR~\cite{kim2016VDSR}. Recently, RCAN~\cite{zhang2018RCAN} employed more than four hundred convolutional layers to enhance image quality. The current batch of CNNs~\cite{ahn2018CARN,tai2017DRRN} are incorporating more convolutional layers to construct deeper networks to improve the image quality and numbers, and this trend has continuously remained a dominant one in deep SR since the inception of SRCNN. 

\subsection{Skip Connections}
Overall, skip connections have played a vital role in the improvement of SR results. These connections can be broadly categorized into four main types: \emph{global} connections, \emph{local} connections, \emph{recursive} connections, and \emph{dense} connections. Initially, VDSR~\cite{kim2016VDSR} utilized global residual learning (GRL) and has shown enormous performance improvement over SRCNN~\cite{dong2016SRCNNPAMI}.  Further, DRRN~\cite{tai2017DRRN} and DRCN~\cite{kim2016DRCN} have demonstrated the effectiveness of recursive connections. Recently,  EDSR~\cite{lim2017EDSR} and RCAN~\cite{zhang2018RCAN} employed local residual learning (LRL) \ie local connections while keeping the global residual learning (GRL) as well. Similarly, RDN~\cite{zhang2018RDN} and ESRGAN~\cite{wang2018esrgan} engaged dense connections and global ones.  Modern CNNs are innovating ways to improve and introduce other types of connections between different layers or modules. In Table~\ref{table:parameters}, we show the skip connections along with the corresponding methods.

\section{Super-resolution competitions}
Recently, the primary reason for the fast-paced research in single-image super-resolution originates from the competitions arranged by companies as well as conferences. Two important challenges are listed below.

\subsection{NTIRE}
To benchmark, the single-image super-resolution, NTIRE\footnote{http://www.vision.ee.ethz.ch/ntire17/} (New Trends in Image Restoration and Enhancement)~\cite{timofte2017ntireFlicker2K} challenge was introduced in 2017. The dataset employed for training and testing is named DIVerse 2K (DIV2K). The challenge has two tracks for evaluating the participants. Track-1, where the classical bicubic degradation is used, and Track-2, where the downsampling is unknown. In the Track-2, the downsampling operator is only known through training LR and HR pair. Furthermore, only blur and decimation is employed with no addition of any noise. The images in the challenge are downscaled using the factors of 2,3 and 4. The aim of this challenge is multi-purpose,
 
\begin{itemize}
\item To introduce a new dataset (DIV2K)
\item To advance the state-of-the-art in super-resolution
\item To compare diverse algorithms
\item To apply challenging settings
\end{itemize}

The NITRE challenge is now extended to more low-level tasks held in conjunction with the computer vision and pattern recognition (CVPR) every year.

\subsection{PIRM}
The next challenge for super-resolution is the Perceptual Image Restoration and Manipulation\footnote{https://www.pirm2018.org/PIRM-SR.html} (PIRM)~\cite{blau20182018PIRM}. This challenge focuses on perceptual quality of the images and quantifies PSNR accuracy jointly. Hence, providing an opportunity to perceptual driven algorithms to advance alongside PSNR targeted algorithms. 

The PIRM challenge employs 4$\times$ factor to test the algorithms competing. The images are downsampled using bicubic kernel degradation. The challenge evaluation is based on traditional full-reference metrics such as PSNR, SSIM, RMSE, FC~\cite{sheikh2005IFC}, LPIPS~\cite{zhang2018LPIPS}, as well as the no-reference methods by Ma~\etal~\cite{ma2017Metric}, NIQE~\cite{mittal2012NIQE}, BRISQUE~\cite{mittal2012BRISQUE}. The perceptual index is computed from Ma~\etal and NIQE~\cite{mittal2012NIQE}.

One hundred images of two sets evaluate the methods. The sets are composed of very diverse contents \eg objects, pedestrians, plants \etc At the time of the competition; the ground-truth high-resolution images are not available to the participants. The authors submit their super-resolved images to an online web portal.  Furthermore, the participants chose datasets for model training.  The PIRM challenge workshop is held in European Conference on Computer Vision (ECCV).

\section{Future Directions/Open Problems}
Although deep networks have shown exceptional performance on the super-resolution task, there remain several open research questions. We outline some of these future research directions below.

\noindent
\textbf{Incorporation of Priors:} Current deep networks for SR are data driven models that are learned in an end-to-end fashion. While this approach has shown excellent results in general, it proves to be sub-optimal when a particular class of degradation occurs for which large amount of training data is non-existent (\eg, in medical imaging). In such cases, if the information about the sensor, imaged object/scene and acquisition conditions is known, useful priors can be designed to obtain high-resolution images. Recent works focusing on this direction have proposed both deep network \cite{ulyanov2018deep} and sparse coding \cite{dong2018learning} based priors  for better super-resolution. 

\noindent
\textbf{Objective Functions and Metrics:} Existing SR approaches predominantly use pixel-level error measures \eg, $\ell_1$ and $\ell_2$  distances or a combination of both \cite{timofte2018ntire}. Since, these measures only encapsulate local pixel-level information, the resulting images do not always provide perceptually sound results. As an example, it has been shown that images with high PSNR and SSIM values give overly smooth images with low perceptual quality \cite{blau20182018}. To counter this issue, several perceptual loss measures have been proposed in the literature. The conventional perceptual metrics were fixed \eg, SSIM \cite{Wang2004}, multi-scale SSIM \cite{Wang2003}, while more recent ones are learned to model human perception of images \eg, LPIPS \cite{Zhang2018} and PieAPP \cite{Prashnani2018}. Each of these measures have their own failure cases. As a result, there is no universal perceptual metric that optimally works in all conditions and perfectly quantifies the image quality. Therefore, the development of new objective functions is an open research problem. To encourage the development in this area, a dedicated challenge and workshop has been organized for perceptually sound image super-resolution approaches (PIRM 2018) \cite{blau20182018}.

\noindent
\textbf{Need for Unified Solutions:} Two or more degradations often happen simultaneously in real life situations. An important consideration in such cases is how to jointly recover images with higher resolution, low noise and enhanced details. Current models developed for SR are generally restricted to only one case and suffer in the presence of other degradations. Furthermore, problems specific models differ in their architectures, loss functions and training details. It is a challenge to design unified models that perform well for several low-level vision tasks, simultaneously \cite{zhang2018SRMDNF}. 

\noindent
\textbf{Unsupervised Image SR:} 
Models discussed in this survey generally consider LR-HR image pairs to learn a super-resolution mapping function. One interesting direction is to explore how SR can be performed for cases where corresponding HR images are not available. One solution to this problem is Zero-shot SR \cite{shocher2017ZSSR} which learns the SR model on a further downsampled version of a given image. However, when an input image is already of poor resolution, this solution cannot work. The unsupervised image SR aims to solve this problem by learning a function from unpaired LR-HR image sets \cite{yuan2018unsupervised}. Such a capability is very useful for real-life settings since it is not trivial to obtain matched HR images in several cases. 

\noindent
\textbf{Higher SR rates:} Current SR models generally do not tackle extreme super-resolution which can be useful for cases such as super-resolving faces in crowd scenes. Very few works target SR rates higher than 8$\times$ (\eg, 16$\times$ and 32$\times$) \cite{lai2017LapSRN}. In such extreme upsampling conditions, it becomes challenging to preserve accurate local details in the image. Further, an open question is how to preserve high perceptual quality in these super-resolved images. 

\noindent
\textbf{Arbitrary SR rates:} In practical scenarios, it's often not known which upsampling factor is the optimal one for a given input. When the downsampling factor is not known for all the images in the dataset, it becomes a significant challenge during training since it becomes hard for a single model to encapsulate several levels of details. In such cases, it is important to first characterize the level of degradation before training and performing inference through a specified SR model. 

\noindent
\textbf{Real vs Artificial Degradation:}
Existing SR works mostly use a bicubic interpolation to generate LR images. Actual LR images that are encountered in real-world scenarios have a totally different distribution compared to the ones generated synthetically using bicubic interpolation. As a result, SR networks trained on artificially created degradations do not generalize well to actual LR images in practical scenarios. One recent effort towards the solution of this problem first learns a GAN to model the real-world degradation \cite{bulat2018learn}. Another recent effort proposes to enrich features by preserving the original spatial resolution and exchanging multi-scale information along the feature processing path for real image SR \cite{zamir2020MIRNet}. Recently, an extensive challenge was organized for real-image super-resolution in CVPR'19 to promote development on this crucial research problem \cite{cai2019ntire, nah2019ntire}.

\section{Conclusions}
Single-image super-resolution is a challenging research problem with important real-life applications. The phenomenal success of deep learning approaches  has resulted in rapid growth in deep convolutional network based techniques for image super-resolution. A diverse set of approaches have been proposed with exciting innovations in network architectures and learning methodologies. This survey provides a comprehensive analysis of existing deep-learning based methods for super-resolution. Through extensive quantitative and qualitative comparisons, we note the following trends in the existing art: (a) GAN-based approaches generally deliver visually pleasing outputs while the reconstruction error based methods more accurately preserve spatial details in an image, (b) for the case of high magnification rates ($8\times$ or above), the existing models generally deliver sub-optimal results, (c) the top-performing methods generally have a higher computational complexity and are deeper than their counterparts, (d) residual learning has been major contributing factor for performance improvement due to its signal decomposition that makes the learning task easier. Overall, we note that the super-resolution performance has been greatly enhanced in recent years with a corresponding increase in the network complexity. Remarkably, the state-of-the-art approaches still suffer from limitations  that restrict their application to key real-world scenarios (\eg, inadequate metrics, high model complexity, inability to handle real-life degradations). We hope this survey will attract new efforts towards the solution of these crucial problems. 

\ifCLASSOPTIONcaptionsoff
  \newpage
\fi

\bibliographystyle{IEEEtran}
\nocite{*}
\bibliography{ref}

\begin{IEEEbiographynophoto}{Saeed Anwar}
is a Research Scientist in the CSIRO (Data61), Australia and adjunct lecturer in Australian National University. He has received his Ph.D. degree from the Australian National University in 2019, the master degree in Erasmus Mundus Vision and Robotics (Vibot), jointly offered by the Heriot-Watt University, United Kingdom, the University of Girona, Spain and the University of Burgundy France in 2012 with distinction. He earned the bachelor degree in Computer Systems Engineering with distinction as well from University of Engineering and Technology (UET), Pakistan, in 2008. During his masters, he carried out his thesis at Toshiba Medical Visualization Systems Europe (TMVSE), Scotland. He has also been a visiting research fellow at Pal Robotics, Barcelona in 2011. He has also worked as a Lecturer and Assistant Professor at the National University of Computer and Emerging Sciences and taught courses at the University of Canberra and the Australian National University. His primary research interests are the low-level vision, image classification, scene understanding, machine learning, computer vision, and optimization.
\end{IEEEbiographynophoto}


\begin{IEEEbiographynophoto}{Salman Khan}
received the B.E. degree in electrical engineering from the National University of Sciences and Technology, Pakistan, in 2012, and the Ph.D. degree from The University of Western Australia, in 2016. His Ph.D. thesis received an honorable mention on the Dean’s List Award. He was a Visiting Researcher with National ICT Australia, CRL, in 2015. He was a Research Scientist with Data61, Commonwealth Scientific and Industrial Research Organization during 2016-2018. He has been a Senior Scientist with Inception Institute of Artificial Intelligence, since 2018, and an Adjunct Lecturer with Australian National University, since 2016. 
His research interests include computer vision, pattern recognition, and machine learning. He was a recipient of several prestigious scholarships, including Fulbright and IPRS. He has served as a program committee member for several premier conferences, including CVPR, ICCV, and ECCV.
\end{IEEEbiographynophoto}

\begin{IEEEbiographynophoto}{Nick Barnes}
received the B.Sc. (Hons.) and Ph.D. degrees in computer vision for robot guidance from The University of Melbourne, Australia, in 1992 and 1999, respectively. In 1999, he was a Visiting Research Fellow with the LIRA Laboratory, University of Genova, Italy. From 2000 to 2003, he was a Lecturer with the Department of Computer Science and Software Engineering, The University of Melbourne. Since 2003, he has been with the  NICTA’s Canberra Research Laboratory, which as become Data61, Commonwealth Scientific and Industrial Research Organization, Australia, where he is currently a Senior Principal Researcher and leads Computer Vision. He has also been an Associate Professor at the Australian National University. His current research interests include dense estimation tasks in computer vision, as well as prosthetic vision, biologically inspired vision, and vision for vehicle guidance. He has published more than 140 research papers on these topics. 
\end{IEEEbiographynophoto}


\vfill


\end{document}